\documentclass[10pt,twocolumn,letterpaper]{article}

\usepackage{cvpr}              % To produce the CAMERA-READY version
\usepackage{graphicx}
\usepackage{amsmath}
\usepackage{amssymb}
\usepackage{booktabs}
\usepackage{times}
\usepackage{epsfig}
\usepackage{multirow}
\usepackage{bbm}
\usepackage[font=small]{caption}
\usepackage{subcaption}
\usepackage{xcolor}

\usepackage[pagebackref,breaklinks,colorlinks]{hyperref}

\usepackage[capitalize]{cleveref}
\crefname{section}{Sec.}{Secs.}
\Crefname{section}{Section}{Sections}
\Crefname{table}{Table}{Tables}
\crefname{table}{Tab.}{Tabs.}

\def\cvprPaperID{****} % *** Enter the CVPR Paper ID here
\def\confName{CVPR}
\def\confYear{2022}

\begin{document}

%%%%%%%%% TITLE
\title{Multi-modal 3D Human Pose Estimation with 2D Weak Supervision in Autonomous Driving}

\author{Jingxiao Zheng\textsuperscript{1}
\and
Xinwei Shi\textsuperscript{1}

\and
Alexander Gorban\textsuperscript{1}

\and
Junhua Mao\textsuperscript{1}

\and
Yang Song\textsuperscript{1}

\and
Charles R. Qi\textsuperscript{1}

\and
Ting Liu\textsuperscript{2}

\and
Visesh Chari\textsuperscript{1}

\and
Andre Cornman\textsuperscript{1}

\and
Yin Zhou\textsuperscript{1}

\and
Congcong Li\textsuperscript{1} \quad Dragomir Anguelov\textsuperscript{1}

\\
\textsuperscript{1} Waymo LLC \quad\textsuperscript{2} Google Research\\
{\tt\small \{jingxiaozheng, xinweis, gorban, junhuamao, yangsong, rqi\}@waymo.com}, \\ {\tt\small liuti@google.com, \{visesh, cornman, yinzhou, congcongli, dragomir\}@waymo.com}
}

\maketitle

%%%%%%%%% ABSTRACT
\begin{abstract}
3D human pose estimation (HPE) in autonomous vehicles (AV) differs from other use cases in many factors, including the 3D resolution and range of data, absence of dense depth maps, failure modes for LiDAR, relative location between the camera and LiDAR, and a high bar for estimation accuracy. 
%Many factors, including the 3D resolution and range, absence of dense depth maps, failure modes for LiDAR, relative location between the the camera and LiDAR, and need for pose estimation accuracy differ between AV and typical indoor applications such as virtual reality, gaming, and animation.
Data collected for other use cases (such as virtual reality, gaming, and animation) may therefore not be usable for AV applications. This necessitates the collection and annotation of a large amount of 3D data for HPE in AV, which is time-consuming and expensive.

In this paper, we propose one of the first approaches to alleviate this problem in the AV setting. Specifically, we propose a multi-modal approach which %takes 2D RGB images and 3D LiDAR point clouds as input, but
uses 2D labels on RGB images as weak supervision to perform 3D HPE. %Our approach also extends the popular PointNet model\cite{pointnet} to a multi-modal setting, and adds an auxiliary segmentation branch.
The proposed multi-modal architecture incorporates LiDAR and camera inputs with an auxiliary segmentation branch. On the Waymo Open Dataset \cite{open_dataset}, our approach achieves a $\sim22\%$ relative improvement over camera-only 2D HPE baseline, and $\sim6\%$ improvement over LiDAR-only model. %These results also translate well to our internal AV dataset which has over $40,000$ LiDAR point clouds in the test set.
Finally, careful ablation studies and parts based analysis illustrate the advantages of each of our contributions.
\end{abstract}

%%%%%%%%% BODY TEXT
\section{Introduction} \label{sec:intro}

3D Human Pose Estimation (3D HPE) for autonomous vehicles (AV) has received little attention in the academic community relative to other applications like animation, games, virtual reality (VR), or surveillance~\cite{HPESurvey} despite its central role in AV.  
Arguably, this could be because 3D HPE in AV differs greatly from HPE in other scenarios. For one, AV requires HPE in outdoor environments and in 3D, which is not the case for animation or games which are not outdoor \cite{Willett:2020:PPS,posetween} or surveillance which is not necessarily in 3D \cite{das2020vpn}. 
Secondly, sensor characteristics and placements for LiDAR follow different logic compared to other depth sensors like in games or VR \cite{weng2018photo,vid2player}. 
Thirdly, requirements for accuracy, real-time prediction and generalization over a wide variety of scenarios are also different.  Animation, %(not real-time)
surveillance, games and VR have relatively lower bars for accuracy compared to AV where HPE is a critical component for the perception module.
% where HPE is a critical component for the safety of passengers, pedestrians and other road users.

\begin{figure}[!t]
\includegraphics[width=\linewidth]{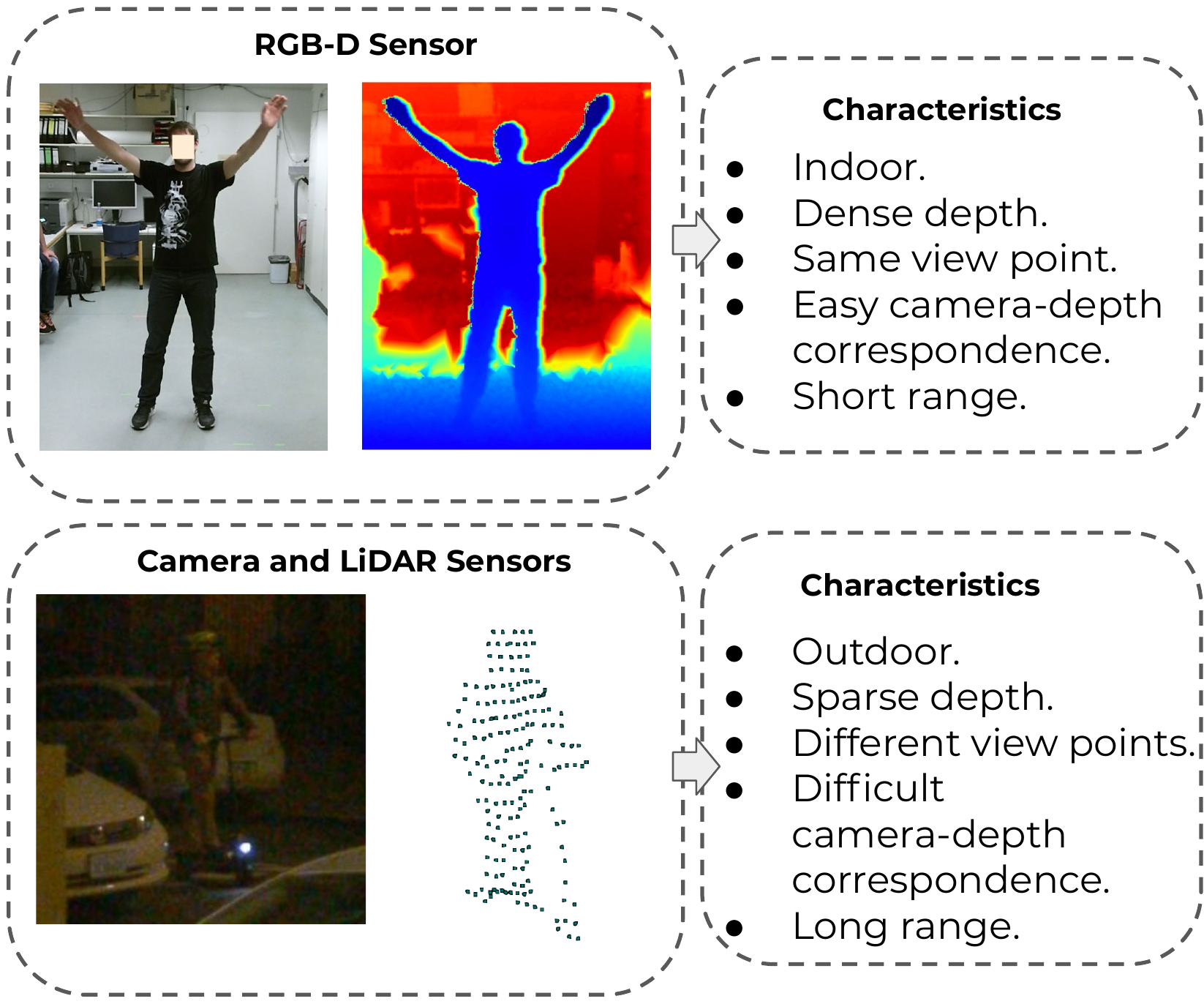}
\caption{Different characteristics of RGB-D and Camera+LiDAR sensors. Top row examples are from dataset \cite{rgbd_voxel}; bottom row examples are from the Waymo Open Dataset \cite{open_dataset}.}
% \caption{The proposed method uses a multi-modal framework that directly predicts 3D keypoints from mutually complementary 2D camera images and 3D point clouds. While 3D labels are expensive to collect, pseudo label generation enables model training on the weak supervision from pure 2D keypoint labels.}
\label{fig:introduction}
\end{figure}

Diving deeper into the sensor, LiDAR differs from other depth sensors in several ways. Figure~\ref{fig:introduction} summarizes these differences and gives visual illustrations. Firstly, LiDAR has longer range and larger FOV than RGB-D sensors, and it is more suitable for outdoor scenes. Point clouds from LiDAR are sparser and sweep a wider range of the environment. %(citations). 
Secondly, LiDARs and cameras may not be co-located on AV platforms. Accurate registration is needed for correspondence between point clouds and image textures.
%(citations). 
Finally, failure cases for LiDAR caused by reflective materials, weather conditions, and dust on sensors differ from other sensor failures due to the difference in the physics of sensing as well as environmental factors. %(citations). 

Given the aforementioned differences and the evidence of 3D HPE models not generalizing across different datasets~\cite{HPESurvey,ws_3d,3d_adversarial,repnet} because of dataset bias, we see the need for developing approaches specific to AV that tackle the problem of 3D HPE. 
%Since existing datasets are mainly captured in constrained scenes \cite{HPESurvey}, 
One straightforward way  to tackle this problem 
would be to collect 3D human pose annotations for a large and diverse dataset of LiDAR point clouds in AV scenarios like the Waymo Open Dataset \cite{open_dataset}. However, the "in-the-wild" setting of 3D HPE for AV presents serious challenges to annotating training data at this scale, in terms of time, cost and coverage of long tail scenarios. 

\begin{figure*}[t]
\centering
\includegraphics[width=\linewidth]{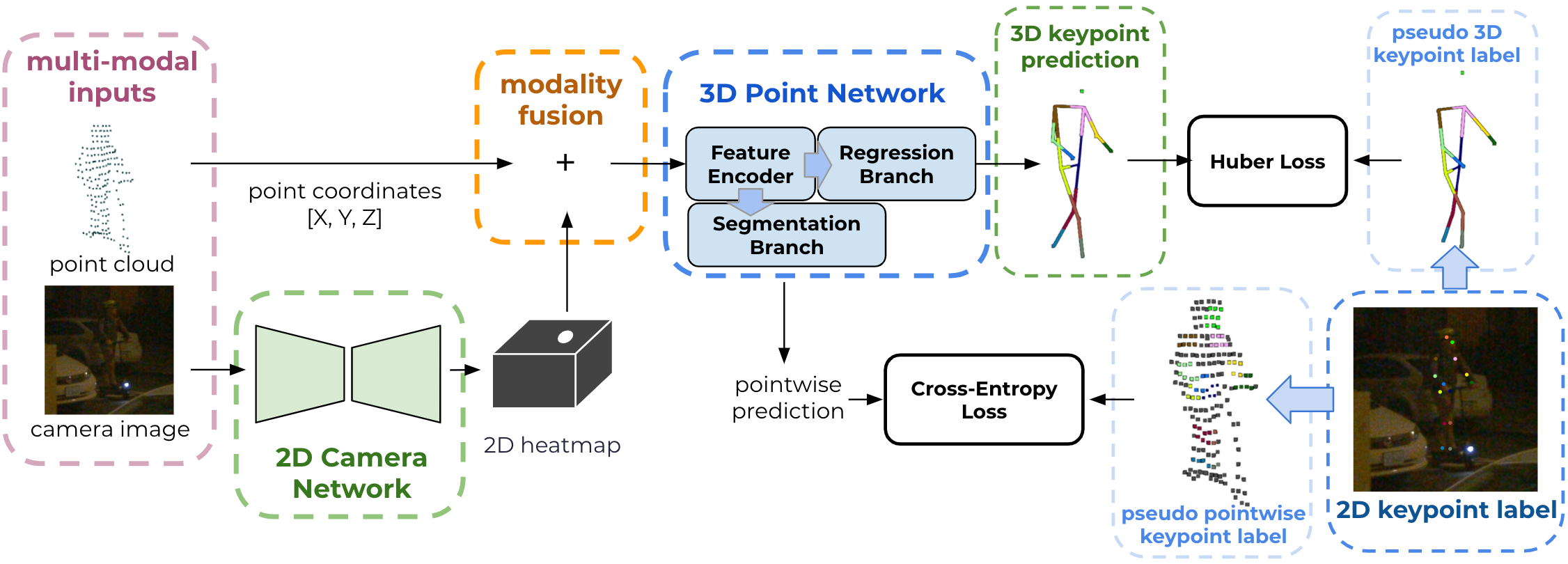}
\caption{Model overview: the model is a cascade of camera network and point network. The camera network takes the 2D camera image as input and predicts the 2D keypoint heatmap. This 2D heatmap is augmented with the point cloud using modality fusion (Figure~\ref{fig:modality_fusion}) and is fed into the point network. The regression branch of the point network predicts 3D keypoint coordinates as output. The auxiliary segmentation branch generates pointwise predictions which are only used for training. The model is trained on pseudo 3D labels and pointwise labels generated from 2D keypoint labels (Figure~\ref{fig:label_generation}).}
\label{fig:system}
\end{figure*}

In this paper, we propose an approach to use widely available and easier to get 2D human pose annotations to drive 3D HPE in a weakly-supervised setting. While the weakly-supervised setting is not uncommon for 3D HPE~\cite{ws_3d}, using LiDAR in the AV setting requires separate consideration for the reasons mentioned thus far. 
Figure~\ref{fig:system} shows the idea of the proposed method. While we use PointNet \cite{pointnet}-inspired architecture as the main point cloud processing network, we cannot fuse camera and LiDAR imagery at the lower levels like in other settings~\cite{earlyfusion} because of the sparsity of LiDAR. 
We propose a cascade architecture with a CNN-based camera network for 2D pose estimation. %, chosen among other approaches based on ablation studies (see supplementary). 
In addition, we add an auxiliary segmentation branch in the point network to introduce stronger supervision to each point via multi-task learning. This gives us an advantage in the "in-the-wild" settings, as shown by the results on the Waymo Open Dataset (Table~\ref{tab:basic} and Table~\ref{tab:ablation}). In the rest of the paper, we show that pose estimation performance benefits from all these designs.

The main contributions of this paper are as follows:
\begin{itemize}
\item We propose a multi-modal framework which fuses RGB camera images and LiDAR point clouds to exploit the texture information and geometry information for 3D pose estimation in challenging AV scenarios.
\item We train 3D pose estimation models by weak supervision from pure 2D labels, which makes the labeling stage much less expensive.
\item We introduce an auxiliary segmentation branch into the point network to improve 3D pose estimation performance via multi-task learning.
%\item We propose a multi-modal framework which fuses the information from camera images with point clouds to improve 3D pose estimation performance.

\end{itemize}

We review related work  in Section~\ref{sec:related}, and follow it up with details about our approach in Section~\ref{sec:method}. Section~\ref{sec:exp} discusses detailed experiments with results on two large datasets, followed by ablation studies and performance analysis (refer to supplementary for additional results).
% other performance analysis parameters like inference time (refer supplementary for additional results). 
Finally, we conclude in Section~\ref{sec:conclusion} with a discussion of avenues for improvement and future directions.

\section{Related Work}\label{sec:related}

In recent years, many methods have been introduced for 3D HPE~\cite{HPESurvey}, although hardly any work has addressed the AV scenario. Most take RGB or RGB-D images as inputs, and operate in monocular, multiview or video settings. 

%For fully-supervised methods,
Monocular 3D HPE approaches like Tome \emph{et al.} \cite{LiftFromDeep} take the simplest of inputs (monocular RGB images) and predict 3D keypoints using a multi-stage method. This classical approach of ``lifting'' 3D keypoints from 2D images has been recently done using deep learning\cite{2d_3dbaseline}, and in the past using a database of 3D skeletons \cite{2dlift_black, 2dlift_sheikh, 2dlift_gao}. Recent criticisms of this approach have focused on over-reliance on the underlying 2D estimator, and of generalization problems\cite{Arnab_2019,HPESurvey}. Extending this approach temporally \cite{3docclusion,2dlift_daniilidis,Arnab_2019,ws_3d} also has been attempted, but still underperforms approaches which use depth information (see \cite{adafuse}, \cite{HPESurvey} table 11). 

Depth based approaches also come in different flavors. Some, like Zimmermann \emph{et al.}, \cite{rgbd_voxel} use a VoxelNet based method on RGB-D images with 3D labels. 
Others might only use point clouds \cite{fuse_2d_3d}, add temporal consistency formulations \cite{temporal_3d}, use a split and recombine approach~\cite{SRNet}, or generate large amounts of synthetic data followed by supervised learning strategy~\cite{cascade_monocular}. Semi-supervised approaches~\cite{ordinaldepth,geometricaware,videopose3d,multiview_3d} have also been recently attempted to deal with the long tail and "in-the-wild" scenarios.
% Cheng \emph{et al.} \cite{3docclusion,2dlift_daniilidis} proposed a model with temporal network on top of multiscale 2D pose heatmaps on RBG videos. Zimmermann \emph{et al.} \cite{rgbd_voxel} proposed a VoxelNet based method on RGB-D images with 3D labels.  \cite{fuse_2d_3d,temporal_3d,semantic_gc,cascade_monocular,SRNet} are some latest methods for fully-supervised 3D pose estimation. 
% %For semi-supervised approaches,
% Zhou \emph{et al.} \cite{ws_3d} introduced an approach on monocular RGB images.
% \cite{3d_adversarial,repnet} are methods on monocular RGB images with adversarial learning technique. %, trained on a mixture of 2D and 3D labeled data. Both methods use adversarial learning technique to enforce predicted 3D poses to be similar to real poses.
% Sun \emph{et al.} \cite{integralpose} introduced a heatmap integral method training on 2D and 3D labels jointly. \cite{ordinaldepth,geometricaware,videopose3d,multiview_3d} are also recent semi-supervised methods. We refer the reader to a comprehensive survey~\cite{HPESurvey} to understand details of various approaches and their tradeoffs.  In the rest of this section, we focus on weakly supervised appraoches and point cloud based approaches that are closer to our work.
% 
\paragraph{Weakly-supervised 3D Human Pose Estimation:}
Besides the above fully- and semi-supervised methods which rely on at least a certain amount of 3D annotations, there are also weakly-supervised methods that use pure 2D annotations. Tripathi \emph{et al.} \cite{posenet3d} introduced a self-supervised method with teacher-student strategy on RGB sequences. %The teacher network predicts a pseudo 3D labels for the student network to fit SMPL models onto. 
Chen \emph{et al.} \cite{geometricself} introduced a weakly-supervised method with cycle GAN\cite{cyclegan}-like structure on pure 2D labels. Other weakly-supervised methods include \cite{3d_eq_2d_matching,multiview_self}. All the above methods are RGB-based, and do not involve the use of point clouds, while our method utilizes point clouds to help to improve the prediction accuracy.

F\"urst \emph{et al.} \cite{hperl} proposed an end-to-end system for 3D detection and HPE for RGB and LiDAR in AV with pure 2D keypoint annotations. However, their work only includes evaluations for 2D HPE and projected 3D HPE, while our approach is evaluated with real 3D annotations.

%There are also some works closely related to 3D human pose estimation, including \cite{view_invariant_embedding} and \cite{sequential_pose_and_shape}.
\paragraph{Point Cloud-Based Approaches:}
Point cloud-based approaches differ from HPE on traditional depth sensors in their ability to handle sparse 3D data~\cite{PCSurvey}.
PointNet \cite{pointnet} is a popular network for point cloud-based classification and segmentation, improved with hierarchical structures in \cite{pointnetplusplus} and  utilized for 3D object detection on RGB-D~\cite{pointnet_detection}, and  hand pose estimation~ \cite{hand_pointnet,hand_offset_pointnet}.
Finally, Zhang \emph{et al.} \cite{weakly3d} proposed a weakly supervised point cloud-based method for 3D human pose estimation. However, their method requires 3D annotations and is only evaluated in indoor RGB-D datasets, while our method works on uncontrolled AV scenarios with pure 2D annotations.

%Compared to \cite{weakly3d}, our approach is improved in the following ways: 1) Supervision: \cite{weakly3d} requires 3D labels for training, while our approach can be trained on pure 2D labels. 2) Modality: the point network in \cite{weakly3d} does not utilize the information from RGB images, while our approach uses modality fusion to augment point clouds with rich texture information from RGB images.%But they only evaluated their method on relatively constrained indoor datasets.

%Compared to previous methods, our approach is different in the following ways: 1) Modality: Previous methods usually rely on RGB images, while our method uses PointNet-inspired model on both point clouds and camera images for 3D human pose estimation. Also, many methods acquire extra information from multiview cameras or video sequences, while our method is based on single frame and single camera. 2) Data quality: previous methods are usually trained on larger and more controlled datasets like Human3.6M dataset \cite{h36m_pami} with millions of images captured in a studio. Our method is trained on an unconstrained dataset where data are captured on public streets by sensors on moving vehicles. 3) Supervision: Most of the methods for 3D pose estimation require a certain amount of data with 3D labels, while our method can be trained purely on 2D labels.

\section{Method}\label{sec:method}
% Two main topics: modality and label. Modality: using point cloud has disadvantages. Use camera modal to help point cloud. Use multi-task learning to help point cloud. Label: use pure 2D labels to train a 3D model.
\subsection{Problem Formulation} \label{sec:prob_formulation}
The 3D pose estimation problem can be described as follows. For each human subject in consideration, there are two modalities of data available: the point cloud and a camera image of the person. The point cloud $\mathbf{P} = \begin{bmatrix}\mathbf{p}_1, \cdots,  \mathbf{p}_i, \cdots, \mathbf{p}_N\end{bmatrix}\in \mathbb{R}^{ N \times d}$, consists of $N$ LiDAR points from a single scan with $d$-dimensional features. In this work, $d=3$. The camera image is an $H\times W \times 3$ RGB image. Assuming we have the extrinsics and intrinsics of the LiDAR and camera,  for each point $\mathbf{p}_i$, its 3D world coordinates $\mathbf{x}_i^{(3)}$ in the point cloud coordinate system and 2D coordinates $\mathbf{x}_i^{(2)}$ in the image coordinate system are known. Given these inputs, the goal is to predict 3D coordinates of $K$ pose keypoints $\{\mathbf{y}_k^{(3)}\}_{k=1}^K\in\mathbb{R}^{K\times 3}$ of the corresponding person. Note that LiDAR point clouds are usually sparse and lie on the surface of the object, while ground truth keypoints are defined inside the human body. Therefore, we cannot choose a subset of $\mathbf{P}$ as the 3D pose of the person and approach 3D HPE in AV as a classification problem.

An overview of the proposed approach is shown in Figure~\ref{fig:system}. Our model is a cascade of a camera network and a point network. The camera network takes a 2D image as input and predicts a 2D keypoints heatmap\cite{simple_baseline_2d}. This heatmap is used to augment the point cloud using modality fusion and fed into the point network. Finally, the regression branch of the point network predicts the 3D coordinates of $K$ keypoints. An auxiliary segmentation branch generates pointwise predictions which are only used for training. The model is trained on pseudo 3D labels and pointwise labels generated from 2D labels.

\subsection{Modality Fusion of LiDAR and Camera}\label{sec:camera_network}
%As we discussed, directly predicting 3D keypoints from an unconstrained 2D camera image is a challenging task because a RGB image lacks explicit depth information. Nevertheless, a camera network could still leverage the rich details in RGB images and provide high quality texture information in 2D space. Therefore,
We introduce a 2D camera network with modality fusion to transfer texture information from RGB images to point clouds. Our camera network follows the architecture proposed in \cite{simple_baseline_2d} which consists of a downscale module and an upscale module. The downscale module is a ResNet-50 network %, which reduces the feature map resolution by 32.
and the upscale module consists of three deconvolutional layers. %, which increase the feature map resolution by 8.
A $1\times 1$ convolutional layer with sigmoid activation follows the upscale module and
produces the output heatmap. The network takes an RGB image with size $H\times W\times 3$ as input and generates a keypoint heatmap $\mathbf{H}=\{\mathbf{h}_{m,n}\}_{m=1,n=1}^{H',W'}$ with size $H'\times W'\times K$, where $K$ is the number of keypoints. % and $H' = H/4,\; W'=W/4$.
Each pixel ${h}_{m,n}$ in the heatmap is a $K$ dimensional vector, indicating the likelihood of the corresponding image pixel belonging to each of the $K$ keypoints. %Each voxel in the heatmap indicates the confidence of the corresponding image pixel belonging to the corresponding keypoint.

\begin{figure*}[t]
\centering
\includegraphics[width=0.6\linewidth]{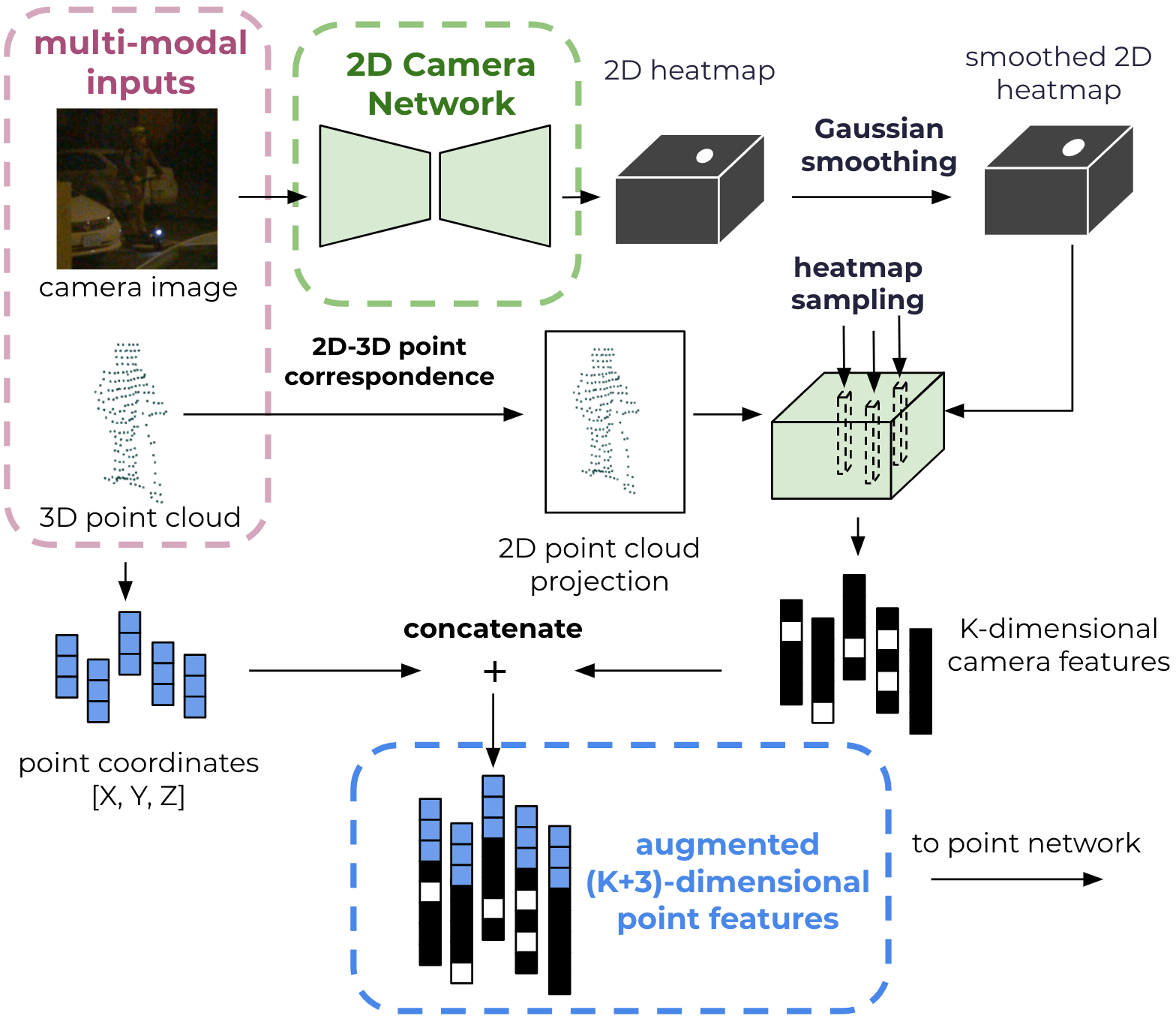}
\caption{Modality fusion: the 2D heatmap from the camera network is first smoothed by Gaussian kernel, then sampled by 2D point cloud projections on the camera image. The sampled heatmap slices are considered as camera features and are concatenated with point coordinates of the point cloud as augmented input to the point network. See Sec~\ref{sec:camera_network} for details.}
\label{fig:modality_fusion}
\end{figure*}
% \subsubsection{Modality Fusion}
The heatmap $\mathbf{H}$ is consequently sampled at points corresponding to the 2D projections on the camera image of 3D LiDAR points, to generate camera features $\mathbf{p}_i^{\text{cam}}$ as shown in Figure~\ref{fig:modality_fusion}. The camera feature for point $i$ is computed as $\mathbf{p}_i^{\text{cam}}=\mathbf{h}_{m(i),n(i)}$,
which is a slice of $\mathbf{H}$ at location $(m(i), n(i))$. Here $m(i) = \mathrm{round}(\frac{W'}{W}x_{1i})$ and $n(i) = \mathrm{round}(\frac{H'}{H}x_{2i})$, where $\mathbf{x}_i^{(2)}=(x_{1i}, x_{2i})$ are the 2D image coordinates of point $i$. %$\mathbf{p}_i^{\text{cam}}$ is a $K$-dimensional vector with sum equal to one. %It could be considered as the keypoint confidences of point $i$ based on the knowledge from the camera image.
In practice, we observe that heatmaps from the camera network are usually very peaky, which contains little information at locations not close to any keypoints. Hence, we apply Gaussian smoothing to enlarge the receptive field at these locations~\cite{heatmap_sample}, so the corresponding point can utilize the information from a larger neighborhood on the image.

Finally, camera features $\mathbf{p}_i^{\text{cam}}$ are concatenated with the original point feature $\mathbf{p}_i$ to generate the augmented point cloud $\mathbf{P}^{\text{aug}}\in\mathbb{R}^{N\times (d+K)}$, which serves as the input of the following point network. This augmentation directly incorporates texture information from RGB images into the point cloud, which helps the LiDAR based point network with information useful for more accurate keypoint predictions. Similar concatenation can be found in \cite{rgbd_voxel}, where voxel representations are concatenated with heatmaps before feeding into a VoxelNet \cite{zhou2017voxelnet}. 

%Based on our experiments%({\color{red}need to find results})
The proposed cascade modality fusion architecture achieves improvements because heatmap predictions from the camera network carry complementary texture related semantic cues that are not present in LiDAR point features. Therefore, augmenting lower-level LiDAR point features with higher-level camera features provides the point network both low- and high- level point cloud information. By introducing modality fusion, we achieve $\sim6\%$ relative improvement on the Waymo Open Dataset compared to the LiDAR-only baseline (Table~\ref{tab:ablation} in Section~\ref{sec:exp}). %On the other hand, if modality fusion is applied in a late fusion manner after the point network producing the global point embedding, it is difficult for the additional camera information to correct mistakes in the point network because low-level LiDAR features are already gone.

\begin{figure*}[t]
\centering
\includegraphics[width=0.6\linewidth]{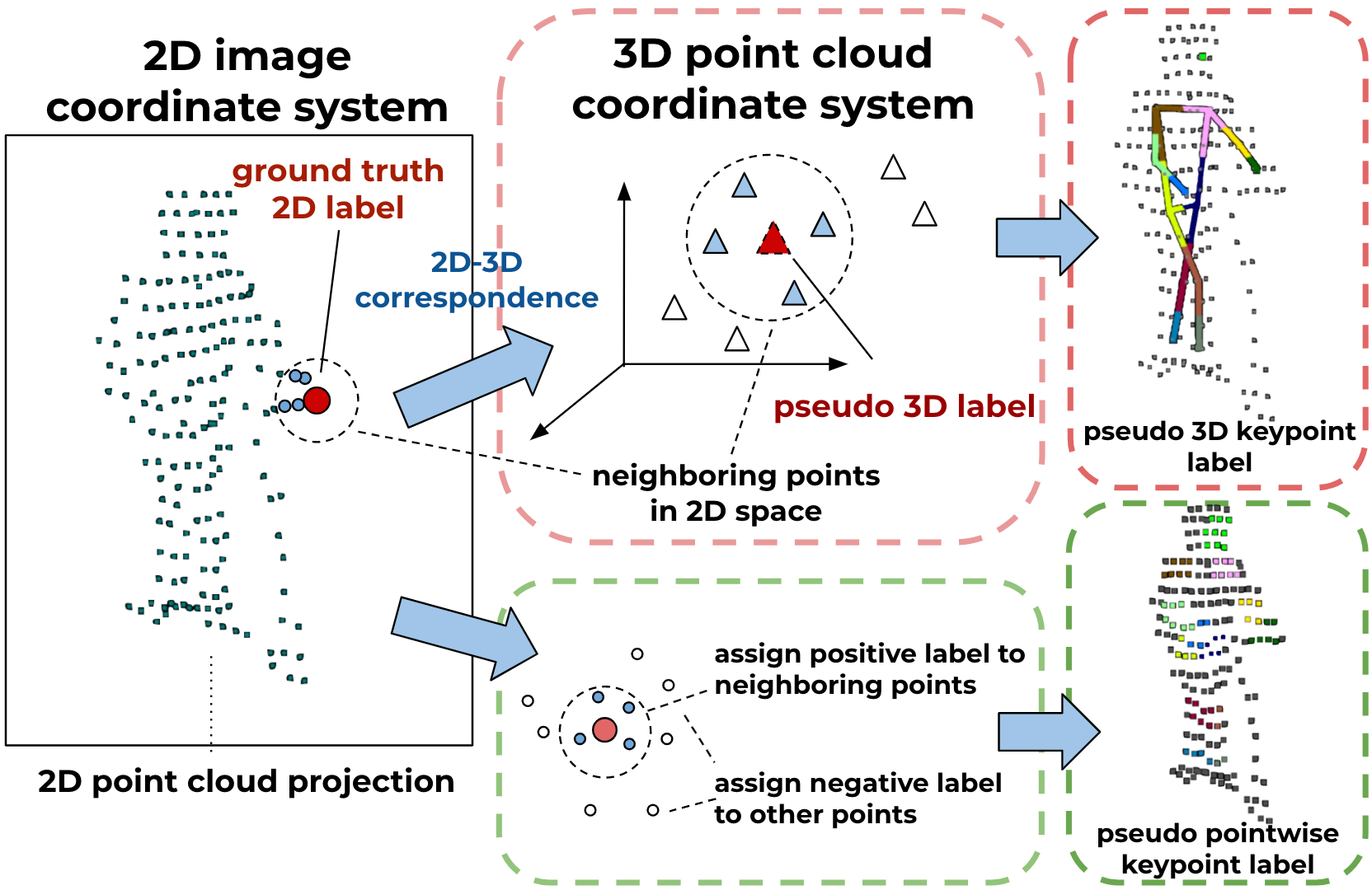}
\caption{Pseudo label generation: a pseudo 3D keypoint label ({\color{red}red} triangle) is computed as the weighed average of 3D coordinates of neighboring points ({\color{blue}blue} triangles and dots) to the keypoint label in 2D space ({\color{red}red} dot). Similarly, to generate pointwise labels, positive labels are assigned to neighboring points ({\color{blue}blue} dots) of a ground truth keypoint ({\color{red}red} dot) in 2D space (best viewed in color). See Sec~\ref{sec:2dto3d} for details.}
\label{fig:label_generation}
\end{figure*}

\subsection{Auxiliary Pointwise Segmentation Branch}\label{sec:point_network}
Our point network is the primary component of the proposed method, which directly generates 3D keypoint prediction from augmented point clouds. %, shown in Figure~\ref{fig:point_net}.
The regression branch predicts a $3K$-dimensional output vector corresponding to the 3D coordinates of $K$ keypoints. 

%In a vanilla PointNet, it is not guaranteed that each point receives enough supervision to train a good feature encoder since the supervision from the training loss is propagated to every point through the bottleneck global feature after max-pooling.

Even though rich camera information from the camera network is provided to the point network by modality fusion, the model's designated output is still a fixed set of keypoints. It is difficult for a global regression loss to guide the point network to effectively utilize the camera information for each point. Therefore, to provide more direct supervision to every individual point, we propose an auxiliary segmentation branch after the feature encoder in the point network, %as shown in Figure~\ref{fig:point_net},
inspired by the architecture of a segmentation PointNet \cite{pointnet}. For each LiDAR point, the segmentation branch predicts the pose keypoint it is closest to. In other words, the segmentation branch generates $N\times K$ confidence scores for assigning $N$ LiDAR points to  $K$ pose keypoints (a point with high score means that it is close to the corresponding keypoint). Here, the keypoint type for each point corresponds to the type of its nearest keypoint.

This additional point-wise loss helps the point network to digest more information from the camera network. By adding the auxiliary segmentation branch and loss, we achieve $\sim1.8\%$ relative improvement on the Waymo Open Dataset compared to the modality-fusion architecture without the segmentation branch (Table~\ref{tab:ablation} in Section~\ref{sec:exp}).

% The point network is trained on both regression task and auxiliary segmentation task in a multi-task manner. The segmentation task introduces point-wise supervision to each point and helps the model to generate better feature encoder, which in turn helps the regression task.
% In practice we would not like the segmentation and regression tasks to compete because of sharing too many weights and deteriorate the performance on the primary regression task. 
% Due to this purpose, instead of using the original design of segmentation PointNet architecture \cite{pointnet} where the segmentation branch is split after the global embedding, we split the segmentation branch from the main branch earlier to introduce more direct supervision. 

\subsection{Weakly-Supervised Model Training}
Training the proposed point network with two branches needs two sets of labels: For the main regression branch, ground truth 3D keypoint coordinates are required; for the segmentation branch, pointwise keypoint type labels are needed. %As we argued in Section~\ref{sec:intro}, collecting 3D keypoint labels for deep network training on unconstrained data is very expensive. Therefore
In the proposed method, we introduce a label generation method to enable model training on pure 2D labels for both tasks.%train our network only on easier-to-collect 2D labels and avoid using any true 3D labels.

%Pseudo label generation: a pseudo 3D keypoint label ({\color{red}red} triangle) is computed as the weighed average of 3D coordinates of neighboring points ({\color{blue}blue} triangles). The blue dots are the neighboring points of a ground truth keypoint (red dot) in 2D image coordinate system, while the blue triangles are their corresponding points in 2D. 
%and dots) to the keypoint label in 2D space ({\color{red}red} dot). 
% Similarly, to generate pointwise labels, positive labels are assigned to neighboring points ({\color{blue}blue} dots) of a ground truth keypoint ({\color{red}red} dot) in 2D space (best viewed in color).

\subsubsection{Label Generation}\label{sec:2dto3d}
As stated in Section \ref{sec:prob_formulation}, we know the 3D coordinates of input points $\{\mathbf{x}_i^{(3)}\}_{i=1}^N$, their corresponding 2D image coordinates $\{\mathbf{x}_i^{(2)}\}_{i=1}^N$, and 2D ground truth keypoints $\{\mathbf{y}_k^{(2)}\}_{k=1}^K$. %, which provides a label generation solution from 2D to 3D, bridged by the correspondence between $\mathbf{x}_i^{(3)}$ and $\mathbf{x}_i^{(2)}$ of the point cloud.
The correspondence is pre-computed by projecting 3D points onto the camera image coordinates according to the camera model. Since the projection is not a one-to-one mapping, directly back-projecting 2D labels to 3D space is impossible.

To generate 3D keypoint labels from the 2D labels and the point cloud, we make the following assumptions:
\begin{enumerate}
    \item the point cloud is dense enough so that there is at least one point in the neighborhood of each keypoint in 2D space;
    \item the human surface is smooth enough so that the depth does not rapidly change in the neighborhood of a keypoint;
    \item point cloud to camera registration is reliable.
\end{enumerate}
Though point clouds will be downsampled to a fixed size before being fed into the point network, pseudo 3D labels are generated based on the point cloud before downsampling. Therefore, the above assumptions hold in most cases. Also, since LiDAR and camera are usually attached to the same rigid object (the vehicle) and are frequently calibrated, it is reasonable to assume that the registration is reliable.

\noindent\textbf{3D Keypoint Coordinates Label Generation:}
Based on our assumption, for each point in the point cloud, its accurate 2D projection on the camera image is known. Therefore, for a ground truth keypoint in 2D coordinates, we can first find its neighboring points in 2D space. Then, based on our assumptions, the depths of these points will be close enough to the true depth of the keypoint. As Figure~\ref{fig:label_generation}, we use the average 3D coordinates of these neighboring points to approximate the coordinates of the keypoint,
\begin{equation}\label{eq:regression_label_generation}
\tilde{\mathbf{y}}_k^{(3)} = \sum_{i=1}^N\alpha_{ik}\mathbf{x}_i^{(3)},\;\alpha_{ik} = \frac{\exp\left(-T \|\mathbf{x}_i^{(2)} - \mathbf{y}_k^{(2)}\|_2^2\right)}{\sum_{j=1}^N\exp\left(-T \|\mathbf{x}_j^{(2)} - \mathbf{y}_k^{(2)}\|_2^2\right)}
\end{equation}
Here $\alpha_{ik}$ weights the contribution of point $i$ to the pseudo keypoint $\hat{\mathbf{y}}_k^{(3)}$ based on their distances to the ground truth keypoint $\mathbf{y}_k^{(2)}$ in 2D space, $T$ is the temperature that controls the softmax operation. %After pseudo 3D labels are generated, the corresponding regression loss can be directly applied in 3D space.

%In the cases that our assumptions are not valid, for example the point cloud is sparse and points are far from the 2D keypoint labels, the approximated 3D labels may not be accurate. Therefore
In case the pseudo 3D labels are not accurate, we also compute the reliability of the 3D approximation for each keypoint as $r_k = \exp\left(-T_r\min_i\|\mathbf{x}_i^{(2)} - \mathbf{y}_k^{(2)}\|_2^2\right)$, where $T_r$ is the temperature factor, to weight the losses on different keypoints during training.%If the distance to the nearest neighbor in 2D space is small, the 3D approximation will be relatively accurate and with higher reliability, and vice versa. 

% 2D-to-3D label generation directly provides depth information to the model which makes the model easier to train. But the approximated labels are sometimes not accurate, especially when the laser points are sparse and far from the keypoints. So we also compute $r_k$, the reliability of the 3D approximation for each keypoint as
% \begin{equation}
% r_k = \exp\left(-T_r\min_i\|\mathbf{x}_i^{(2)} - \mathbf{y}_k^{(2)}\|_2^2\right)
% \end{equation}
% where $T_r$ is the temperature factor. If the distance to the nearest neighbor of a keypoint in 2D space is small, the 3D approximation will be relatively accurate and the reliability score will be high, and vice versa. $r_k$ are used to weight the losses on different keypoints during training.

\noindent\textbf{Pointwise Keypoint Type Label Generation:}
To generate pointwise type labels for the segmentation task, we simply assign all neighboring points of a keypoint in 2D space to the corresponding keypoint type, shown in Figure~\ref{fig:label_generation}. The type label $l_{ik}$ for point $i$ with respect to keypoint $k$ is generated by
\begin{equation}
l_{ik} =
\begin{cases}
	1 & \text{if } \|\mathbf{x}_i^{(2)} - \mathbf{y}_k^{(2)}\|_2 \leq r,\\
    0 & \text{otherwise}
    \end{cases}
\end{equation}
where $r$ is the neighboring radius for positive samples. %Similar operation has been used in \cite{}

With the generated pseudo 3D labels to train the 3D keypoint model, we achieve $\sim22\%$ relative improvement on the Waymo Open Dataset compared to the baseline of predicting 2D keypoints with 2D labels and lifting to 3D (Table~\ref{tab:ablation} in Section~\ref{sec:exp}).

\subsection{Training Losses} \label{sec:training_loss}
\textbf{Point Network:}
The training loss for the regression branch is a Huber loss $L_{\text{reg}}$ on the generated pseudo 3D labels, weighted by the reliability $r_k$. The loss for the segmentation branch is a cross-entropy loss $L_{\text{seg}}$ on the pseudo pointwise labels weighted by different positive/negative sample weights. The overall loss for the point network is
\begin{equation}
L = L_{\text{reg}} + \lambda L_{\text{seg}}
\end{equation}
where $\lambda$ is used to weight the auxiliary segmentation loss.

\noindent \textbf{Camera Network:} Similar to \cite{simple_baseline_2d}, the camera network is trained on a mean-squared-error loss with ground truth 2D heatmap. We train the camera network independently, then freeze it during point network training.

Note that we only train and evaluate on visible keypoints. During training, keypoint losses are only applied on visible keypoints, which means we will not generate pseudo labels for occluded keypoints. In Section~\ref{sec:exp}, we show that even trained on visible keypoints only, the model is able to predict reasonable keypoints for occluded body parts. For more details of training losses, please refer to the supplementary material.

\section{Experiments}\label{sec:exp}
\subsection{Data and Evaluation Metrics} \label{sec:evaluation_metrics}
\paragraph{Training Data:}
We collect an internal dataset with RGB images and LiDAR point clouds similar to the Waymo Open Dataset \cite{open_dataset}. It consists of a total number of 197,381 pedestrians. These pedestrians are labeled with 2D keypoint labels of 13 keypoint types (\emph{nose}, \emph{left/right shoulders}, \emph{left/right elbows}, \emph{left/right wrists}, \emph{left/right hips}, \emph{left/right knees} and \emph{left/right ankles}) in the camera image. These samples are split into a training set with 155,182 pedestrians and a test set with 42,199 pedestrians. The training set with pure 2D labels is used to train the proposed model.

\paragraph{3D Evaluation Data and Metrics:}
The Waymo Open Dataset serves as our 3D evaluation set. It is composed of sensor data collected by Waymo cars under a variety of conditions. It contains 1,950 segments of 20s each, with sensor data including point clouds from LiDAR and RGB images captured by cameras. For 3D evaluation, we labeled 986 pedestrians with 3D keypoint coordinates of 13 keypoint types (same as our internal dataset) on LiDAR point clouds. We are looking to release these labels for evaluation once obtained related approvals.%These labels will be released for evaluation. 

Evaluation results are reported in the OKS (Object Keypoint Similarity) accuracy (OKS/ACC) metric, which is similar to the OKS/AP metric introduced in COCO keypoint challenge \cite{coco} (please refer to the supplementary material for more details), and MPJPE (Mean Per Joint Position Error) \cite{h36m_pami} in 3D coordinates.

\paragraph{2D Evaluation Data and Metrics:}
The test set of our internal dataset serves as the 2D evaluation set. Evaluation results are reported in the OKS/ACC metric in 2D coordinates, after the 3D predictions are projected to 2D space by the corresponding lidar to camera projections.

\paragraph{Labeling:}
For 2D/3D keypoint labeling on the Waymo Open Dataset and the Internal Dataset, we adopt a definition of keypoints similar to the COCO Challenge. Each keypoint is labeled by multiple annotators, whose results are aggregated to determine the final label. For 2D labeling, we only label 2D coordinates of keypoints that are visible in the camera image. For occluded keypoints, we label them as invisible. 3D labeling is similar, where we only label keypoints that are visible from the point clouds. Since we pair each LiDAR with its closest camera in location,
%Empirically, %Since the cameras and the LiDAR are not too far apart,
the occlusion status of keypoints is mostly consistent between 2D and 3D.

% \paragraph{ITOP Dataset}
% The ITOP dataset \cite{itop} is a depth map based 3D human pose dataset. It consists of side-view and top-view depth maps of 20 subjects posing different actions in a studio scene. The dataset provides both 2D and 3D ground truth keypoint coordinates of 15 keypoint types (\emph{head}, \emph{neck}, \emph{left/right shoulders}, \emph{left/right elbows}, \emph{left/right hands}, \emph{torso} \emph{left/right hips}, \emph{left/right knees} and \emph{left/right feet}) and is split into 40K training and 10K testing samples. Point clouds are rendered from depth maps. Evaluation results are reported in PCK (Percentage of Correct Keypoints) \cite{PCK} at 10cm, and MPJPE metrics in 3D coordinates. Only side-view depth maps are used for our training and evaluation. To the best of our knowledge, this dataset is the closest to the setting of the proposed method among public datasets. However, it is still different in that it only has one data source, the depth maps, while the proposed method benefits most from fusing different modalities (RGB camera images and LiDAR point cloud). We conduct experiments on this dataset to compare with recent work on point cloud-based pose estimation. In the experiments, we treat the rendered depth images as camera images since real camera images are not available. 
% \vspace{-5pt}

%Rendered depth images and point clouds are provided according to the depth maps.

\begin{table}[t]
\centering\resizebox{\linewidth}{!}
{\begin{tabular}{|c|c|c|c|}
\hline
\multirow{2}{*}{Methods} & \multicolumn{2}{c|}{Waymo Open Dataset} & Internal Dataset\\ 
\cline{2-4}
& OKS@3D$\uparrow$ & MPJPE$\downarrow$ & OKS@2D$\uparrow$\\
\hline
camera-only \cite{simple_baseline_2d} & 51.74\%& 13.90cm & 78.19\%\\
\hline
LiDAR-only & 59.58\%&10.80cm&77.53\% \\
\hline
multi-modal & \textbf{63.14\%}&\textbf{10.32cm}& \textbf{82.94\%}\\
\hline
\end{tabular}}
\caption{Comparison of camera-only, LiDAR-only, and multi-modal models.  As described in Section~\ref{sec:evaluation_metrics}, OKS@3D stands for OKS/ACC in 3D evaluation, OKS@2D stands for OKS/ACC in 2D evaluation, and MPJPE is another evaluation metric in 3D. These metrics are used throughout the experiments. The proposed multi-modal model achieves the best results on both datasets.}
\label{tab:basic}
\end{table}

%Results on the Waymo Open Dataset and the Internal Dataset

\begin{table*}[t]
\centering\resizebox{0.7\linewidth}{!}
% \tiny
{\begin{tabular}{|c|c|c|c|c|c|c|}
\hline
\multirow{2}{*}{parts} & \multicolumn{2}{c|}{camera-only} & \multicolumn{2}{c|}{LiDAR-only} & \multicolumn{2}{c|}{multi-modal}\\
\cline{2-7}
& OKS@3D & OKS@2D & OKS@3D & OKS@2D & OKS@3D & OKS@2D \\
\hline	
nose &	24.50\% & \textbf{75.10\%} &	23.83\% & 56.27\% &	\textbf{29.74\%} & 72.17\% \\
\hline	
shoulder &	65.41\%&83.38\% &	\textbf{77.04\%} & 85.68\% &	76.93\%&\textbf{87.89\%} \\
\hline	
elbow &	65.61\%&82.63\% &	66.61\%&78.72\% &	\textbf{72.49\%}&\textbf{84.82\%} \\
\hline	
wrist &	45.99\%&79.03\% &	30.37\%&64.10\%&	\textbf{46.97\%}&\textbf{79.17\%} \\
\hline	
hip &	57.69\%&87.97\% &	\textbf{79.42\%}&90.33\% &	74.76\%&\textbf{92.37\%} \\
\hline	
knee &	65.40\%&85.91\% &	77.48\%&86.82\% &	\textbf{78.04\%}&\textbf{90.05\%} \\
\hline	
ankle &	62.68\%&84.17\% &	69.06\%&85.63\% &	\textbf{72.30\%}&\textbf{88.72\%} \\
\hline
\hline
overall & 51.74\% & 78.19\% & 59.58\% & 77.53\% &	\textbf{63.14\%} & \textbf{82.94\%} \\
\hline
\end{tabular}}
\caption{Per-keypoint comparison of camera-only, LiDAR-only, and multi-modal models. OKS@3D is on the Waymo Open Dataset and OKS@2D is on the Internal Dataset. Note that the per-keypoint OKS is computed on each keypoint separately (please refer to supplementary for details). The proposed multi-modal model achieves the best results on most of the keypoint types.} 
\label{tab:per-keypoint}
\end{table*}

\begin{table*}[t]
\centering\resizebox{0.6\linewidth}{!}
{\begin{tabular}{|c c c|c|c|c|}
\hline
  \multicolumn{3}{|c|}{Configurations} & \multicolumn{2}{|c|}{Waymo Open Dataset} & Internal Dataset\\
\hline
Reg. Loss & Seg. Loss & Camera & OKS@3D$\uparrow$ & MPJPE$\downarrow$&
  OKS@2D$\uparrow$\\
\hline
\checkmark& & & 59.10\%&10.93cm&77.52\%\\
\cline{4-6}
\checkmark& \checkmark& &59.58\%&10.80cm&77.53\%\\
\cline{4-6}
\checkmark& &\checkmark &62.03\%&10.53cm&82.51\%\\
\cline{4-6}
\checkmark&\checkmark&\checkmark& \textbf{63.14\%}&\textbf{10.32cm}& \textbf{82.94\%} \\
\hline
\end{tabular}}
\caption{Ablation studies on different model architectures. The best performance is achieved by using multi-modal architecture with auxiliary segmentation loss.}%OKS@3D stands for OKS/ACC in 3D space. OKS@2D stands for OKS/ACC in 2D space. MPJPE is evaluated in 3D space.}
\label{tab:ablation}
\end{table*}

\paragraph{Implementation Details:}
For the Waymo Open Dataset and the Internal Dataset, we resize all camera images to $256\times256$, and randomly sub-sample the input point cloud to a fixed size of 256 points (we did not observe obvious performance gain for larger number of points). %For the ITOP dataset, the pre-processing steps in \cite{weakly3d} are followed.
Please refer to the supplementary material for more training details.

\subsection{Performance Analysis}

To show the effectiveness of the proposed method, we compare with the following models.

\textbf{Camera-only model:} we use the same camera network \cite{simple_baseline_2d} as the proposed method to predict 2D keypoints. Then 2D-to-3D keypoint lifting is implemented by the 2D-to-3D pseudo label generation method introduced in Section~\ref{sec:2dto3d}, the same way as we generate training labels.

\textbf{LiDAR-only model:} we use the proposed point network to predict 3D keypoints without the modality fusion, i.e. only use 3D coordinates of the point clouds as features.

\begin{figure}[t]
\centering
\begin{subfigure}[b]{0.23\textwidth}
        \includegraphics[width=\textwidth]{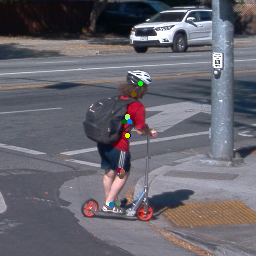}
        \caption{No Camera (LiDAR-Only)} \label{fig:ablation_point_only}
    \end{subfigure}
    ~
    \begin{subfigure}[b]{0.23\textwidth}
        \includegraphics[width=\textwidth]{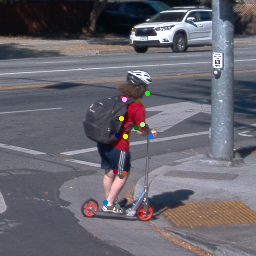}
        \caption{Inception 48x48} \label{fig:ablation_i48}
    \end{subfigure}
    ~ 
    \begin{subfigure}[b]{0.23\textwidth}
        \includegraphics[width=\textwidth]{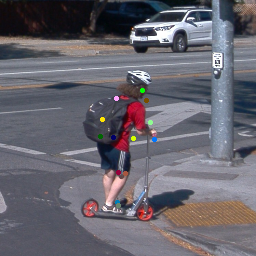}
        \caption{Inception 64x64} \label{fig:ablation_i64}
    \end{subfigure}
    ~ 
    \begin{subfigure}[b]{0.23\textwidth}
        \includegraphics[width=\textwidth]{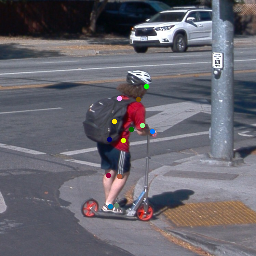}
        \caption{ResNet50 256x256} \label{fig:ablation_r50}
    \end{subfigure}
\caption{3D predictions with different camera image sizes and camera network backbones from the Waymo Open Dataset (best viewed in color). ResNet50 with 256x256 image size predicts the most accurate keypoints.}
\label{fig:ablation}
\end{figure}

% \subsection{Observations}
Experimental results on two datasets are shown in Table~\ref{tab:basic}. % and \ref{tab:itop}.
Table~\ref{tab:per-keypoint} further shows per-keypoint results.
These results
show that our method outperforms all baselines %and recent approaches
in the corresponding datasets. We also have the following observations.

\textbf{Training on pseudo labels is effective.} LiDAR-only baseline and the proposed method both outperform camera-only baseline on 3D metrics on the Waymo Open Dataset. Since the camera-only baseline is also trained on 2D labels and utilizes point clouds to lift the predictions to 3D space, the results indicate that it is more effective to directly train a 3D human pose model on pseudo labels generated from 2D ground truth.

\textbf{Camera image improves 3D prediction.} The proposed method performs better than LiDAR-only baseline on 3D metrics, which demonstrates that the information from 2D camera images helps 3D pose estimation. Table~\ref{tab:per-keypoint} shows that the proposed method outperforms baselines on almost all body parts. Compared to the LiDAR-only baseline, the margins are larger for difficult body parts like \emph{elbows} or \emph{wrists}, which shows that texture information from camera images is especially helpful for keypoints that are hard to localize.

\textbf{Point cloud improves 2D prediction.} LiDAR-only baseline has comparable performance with the camera-only baseline for 2D pose estimation on 2D metrics on the Internal Dataset. The proposed method surpasses the camera-only baseline, even if the models are not directly trained for 2D pose estimation. It shows that the depth information from 3D LiDAR point clouds also improves 2D pose estimation performance.

\textbf{Modality fusion benefits from both modalities.} The proposed method achieves the best performance on all metrics for both datasets. %under weakly-supervised setting.
It proves that camera images and LiDAR point clouds provide complementary information, and modality fusion combines these sources of information to improve the overall performance.

% \begin{table}[t]
% \centering\resizebox{0.5\textwidth}{!}
% {\begin{tabular}{|c|c|c|c|c|c|c|c|c|}
% \hline
%   \multirow{2}{*}{Camera Network} & \multicolumn{8}{|c|}{Per-Keypoint MPJPEs}\\
%   \cline{2-9}
%   & \emph{all} & \emph{elbow} & \emph{wrist} & \emph{hip} & \emph{knee} & \emph{ankle} & \emph{nose} & \emph{shoulder} \\
% \hline
% No Camera & 0.1080 & 0.1006 & 0.1652 & \textbf{0.1081} & 0.0944 & 0.1163 & 0.0814 & 0.0850\\
% \hline
% Inception 48x48 & \textbf{0.1026} & 0.0940 & 0.1501 & 0.1113 & \textbf{0.0896} & \textbf{0.1100} & 0.0762 & \textbf{0.0814}\\
% \hline
% Inception 64x64 & 0.1028 & 0.0931 & 0.1473 & 0.1113 & 0.0910 & 0.1102 & \textbf{0.0760} & 0.0830\\
% \hline
% ResNet50 256x256 & 0.1032 & \textbf{0.0891} & \textbf{0.1320} & 0.1205 & 0.0925 & 0.1107 & 0.0837 & 0.0872\\
% \hline
% \end{tabular}}
% \caption{Per-keypoint performance with different camera networks and image sizes on the Waymo Open Dataset. ResNet50 with 256x256 image size performs the best on challenging keypoints like \emph{elbow} and \emph{wrist} with large margins, but slightly worse than smaller image sizes on other keypoint types.}
% \label{tab:ablation_cam_per_keypoint}
% \end{table}

\begin{figure*}[t]
\centering
\begin{subfigure}[b]{0.23\textwidth}
    \includegraphics[width=\textwidth]{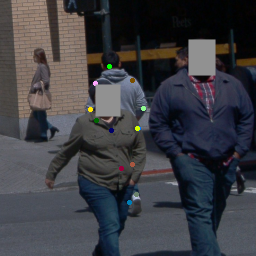}
    \caption{}\label{fig:occ_1}
\end{subfigure}
~
\begin{subfigure}[b]{0.17\textwidth}
    \includegraphics[width=\textwidth]{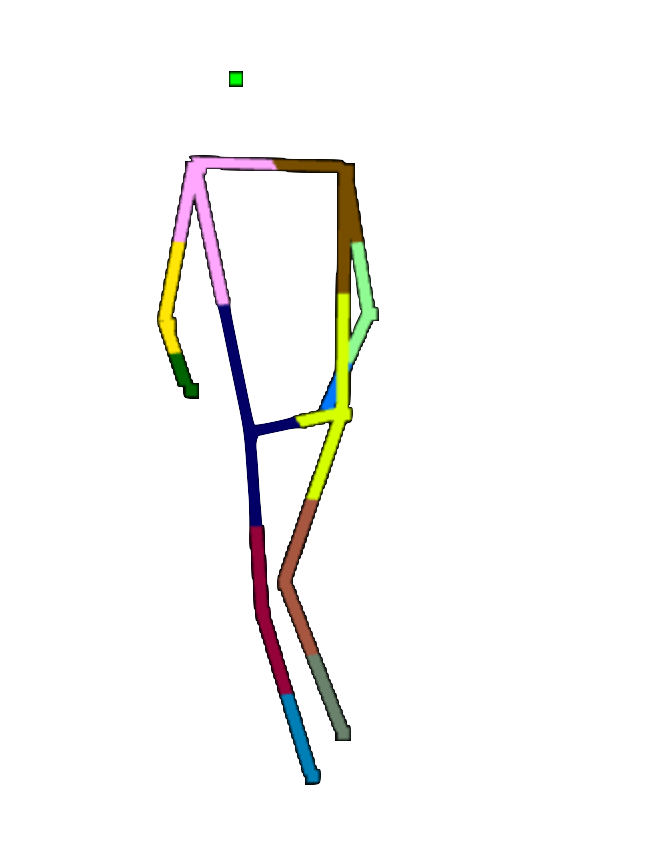}
    \caption{}\label{fig:occ_2}
\end{subfigure}
~
\begin{subfigure}[b]{0.23\textwidth}
    \includegraphics[width=\textwidth]{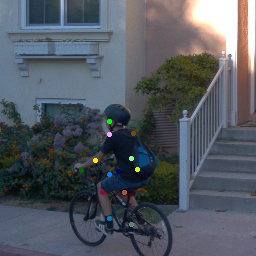}
    \caption{}\label{fig:occ_3}
\end{subfigure}
~
\begin{subfigure}[b]{0.17\textwidth}
    \includegraphics[width=\textwidth]{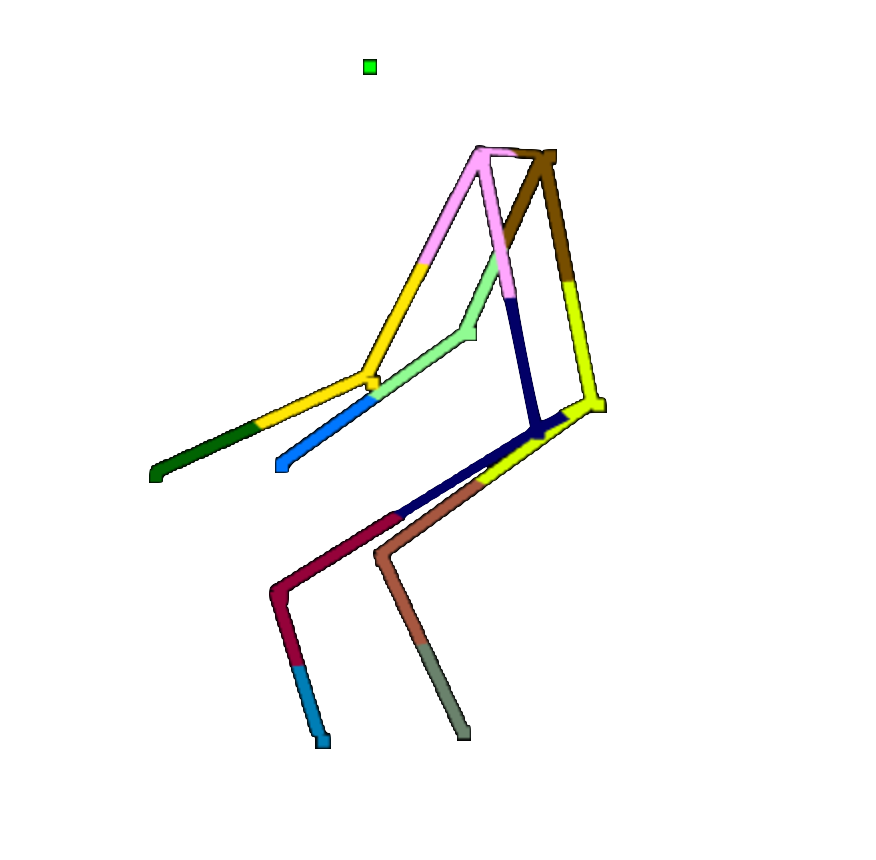}
    \caption{}\label{fig:occ_4}
\end{subfigure}
% ~
% \begin{subfigure}[b]{0.15\textwidth}
%     \includegraphics[width=\textwidth]{figs/new_example_1.png}
%     \caption{}\label{fig:occ_1}
% \end{subfigure}
% ~
% \begin{subfigure}[b]{0.12\textwidth}
%     \includegraphics[width=\textwidth]{figs/new_example_1_3d.png}
%     \caption{}\label{fig:occ_2}
% \end{subfigure}
\\
\begin{subfigure}[b]{0.23\textwidth}
    \includegraphics[width=\textwidth]{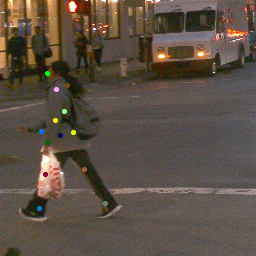}
    \caption{}\label{fig:new_2}
\end{subfigure}
~
\begin{subfigure}[b]{0.17\textwidth}
    \includegraphics[width=\textwidth]{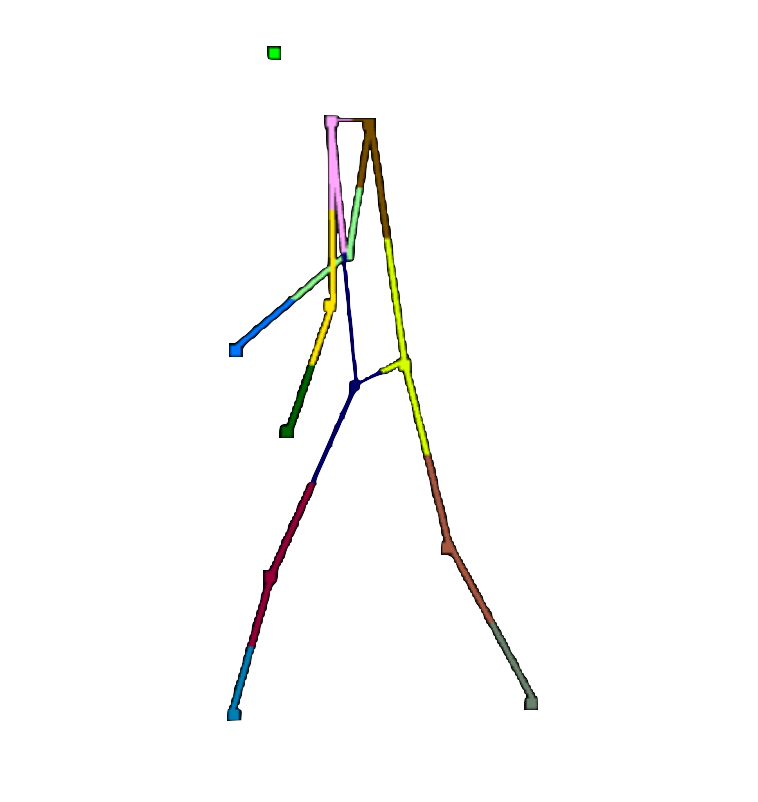}
    \caption{}\label{fig:new_2_3d}
\end{subfigure}
~
\begin{subfigure}[b]{0.23\textwidth}
    \includegraphics[width=\textwidth]{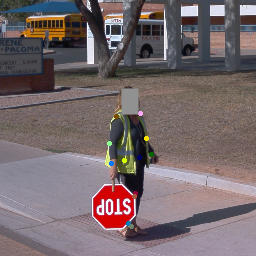}
    \caption{}\label{fig:occ_5}
\end{subfigure}
~
\begin{subfigure}[b]{0.17\textwidth}
    \includegraphics[width=\textwidth]{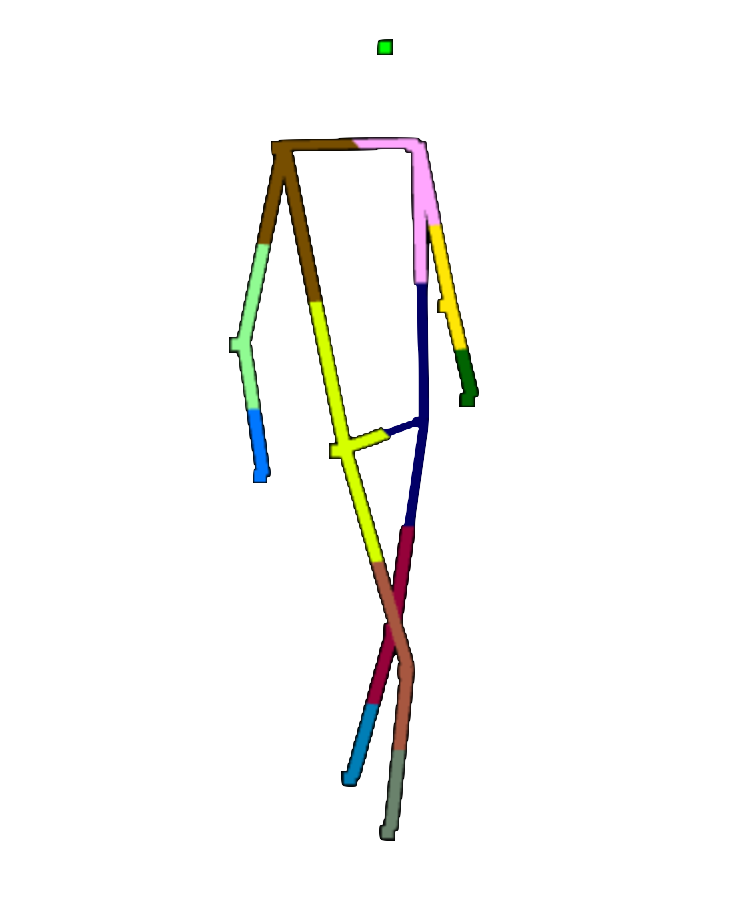}
    \caption{}\label{fig:occ_6}
\end{subfigure}
\\
\begin{subfigure}[b]{0.23\textwidth}
    \includegraphics[width=\textwidth]{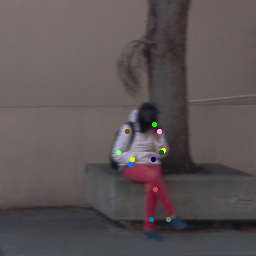}
    \caption{}\label{fig:new_3}
\end{subfigure}
~
\begin{subfigure}[b]{0.17\textwidth}
    \includegraphics[width=\textwidth]{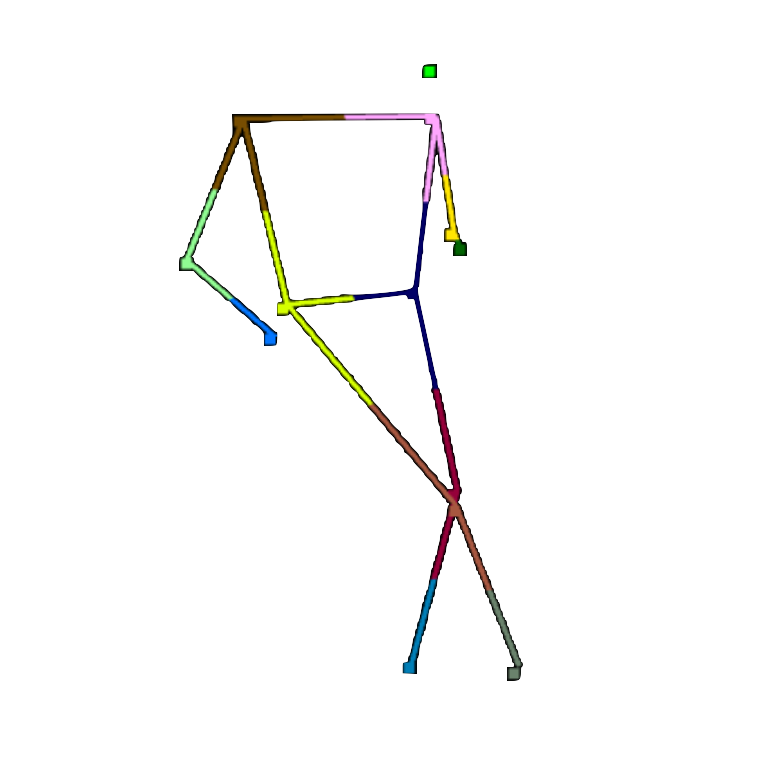}
    \caption{}\label{fig:new_3_3d}
\end{subfigure}
~\begin{subfigure}[b]{0.23\textwidth}
    \includegraphics[width=\textwidth]{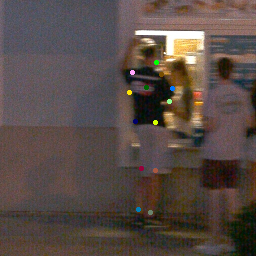}
    \caption{}\label{fig:occ_7}
\end{subfigure}
~
\begin{subfigure}[b]{0.17\textwidth}
    \includegraphics[width=\textwidth]{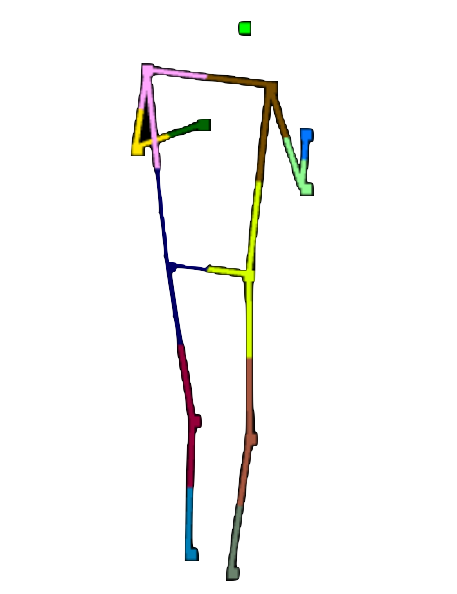}
    \caption{}\label{fig:occ_8}
\end{subfigure}
\caption{Results on the Waymo Open Dataset. \ref{fig:occ_2}, \ref{fig:occ_4}, \ref{fig:occ_6}, \ref{fig:occ_8} are 3D predictions. \ref{fig:occ_1}, \ref{fig:occ_3}, \ref{fig:occ_5}, \ref{fig:occ_7} show the corresponding 2D projections overlaid on camera images (3D predictions may not be shown under the same viewpoint as the camera images. Best viewed in color). More results can be found in supplementary.}
\label{fig:prediction_visualization}
\end{figure*}

Figure~\ref{fig:prediction_visualization} shows some qualitative results of the proposed method on the Waymo Open Dataset. In these examples, pedestrians are either occluded (\ref{fig:occ_1}), in an irregular pose (\ref{fig:occ_3}, \ref{fig:new_3}), or carrying a large object (\ref{fig:occ_5}, \ref{fig:new_2}). The proposed method accurately predicts the visible human keypoints and provides reasonable guesses for the occluded keypoints. Figure~\ref{fig:occ_7} is a failure case where the camera image is blurred because of the sensor motion. It causes an inaccurate prediction of the left wrist. More qualitative results can be found in the supplementary. %Figure~\ref{fig:itop_prediction_visualization} shows 3D predictions from the proposed method under fully-supervised setting on the ITOP dataset, in which our method successfully captures the subject's complicated poses in 3D space.

\subsection{Ablation Studies}
\subsubsection{Ablation Study on Model Architecture}
We conduct ablation studies to further demonstrate the effectiveness of our key designs: the auxiliary segmentation branch and the modality fusion with camera network. The results are shown in Table~\ref{tab:ablation}, where \emph{Reg. Loss} means using regression loss (the primary loss) to train the point network, \emph{Seg. Loss} means auxiliary segmentation branch being added (see Section~\ref{sec:training_loss}), and \emph{Camera} means using modality fusion with camera features. The results show that, by adding key features to the model, the performance improves consistently on all datasets. We also observe that the segmentation branch and modality fusion provide complementary improvements.
% Trying to transpose table 4 
\begin{table*}[t]
\centering\resizebox{0.7\textwidth}{!}
{\begin{tabular}{|c|c|c|c|c|c|}
\hline
    \multirow{10}{*}{\rotatebox{90}{Per-keypoint MPJPEs}}& & \multicolumn{4}{|c|}{Camera Network} \\
    \cline{2-6}
 & & No camera & Inception 48x48 & Inception 64x64 & ResNet50 256x256 \\
    \cline{2-6}
    & \emph{all} & 0.1080 & \textbf{0.1026} &  0.1028& 0.1032\\
    \cline{2-6}
    & \emph{elbow} & 0.1006 & 0.0940& 0.0931 &\textbf{0.0891} \\
    \cline{2-6}
    & \emph{wrist} & 0.1652& 0.1501& 0.1473&\textbf{0.1320} \\
    \cline{2-6}
    & \emph{hip} & \textbf{0.1081}& 0.1113&0.1113 &0.1205 \\
    \cline{2-6}
    & \emph{knee} & 0.0944& \textbf{0.0896}&0.0910 &0.0925 \\
    \cline{2-6}
    & \emph{ankle} & 0.1163& \textbf{0.1100}&0.1102 &0.1107 \\
    \cline{2-6}
    & \emph{nose} & 0.0814& 0.0762& \textbf{0.0760}&0.0837 \\
    \cline{2-6}
    & \emph{shoulder} & 0.0850& \textbf{0.0814}&0.0830 &0.0872 \\
    \hline
\end{tabular}}
\caption{Per-keypoint performance with different camera networks and image sizes on the Waymo Open Dataset. ResNet50 with 256x256 image size performs the best on challenging keypoints like \emph{elbow} and \emph{wrist} with large margins, but slightly worse than smaller image sizes on other keypoint types.}%Notice that they all outperforms camera-only and LiDAR-only baselines.}
\label{tab:ablation_cam_per_keypoint}
\end{table*}

%We speculate that the reason is that the camera network utilizes texture information, and the segmentation branch introduces stronger pointwise supervision. Both of them contribute to learning better point embeddings and predicting more accurate keypoints.

\begin{table*}[t]
\centering\resizebox{0.65\linewidth}{!}
{\begin{tabular}{|c|c c|c|c|c|}
\hline
  \multirow{2}{*}{Camera Network} & \multicolumn{2}{|c|}{Config} &  \multicolumn{2}{|c|}{Waymo Open Dataset} &
  Internal Dataset\\
  \cline{2-6}
  & Reg. &Seg.& OKS@3D$\uparrow$ & MPJPE$\downarrow$ & OKS@2D$\uparrow$ \\
\hline
\multirow{2}{*}{No Camera} &
\checkmark& & 59.10\%&10.93cm&77.52\%\\
\cline{4-6}
&\checkmark& \checkmark&59.58\%&10.80cm&77.53\%\\
\hline
\multirow{2}{*}{Inception 48x48} & \checkmark&&61.12\%&10.51cm&78.72\%\\
\cline{4-6}
& \checkmark&\checkmark&62.22\%&\textbf{10.26cm}&79.55\%\\
\hline
\multirow{2}{*}{Inception 64x64} & \checkmark& &61.05\%&10.46cm&78.95\%\\
\cline{4-6}
 & \checkmark&\checkmark&62.52\%&10.28cm&79.44\%\\
\hline
\multirow{2}{*}{ResNet50 256x256}& \checkmark&&62.03\%&10.53cm&82.51\%\\
\cline{4-6}
& \checkmark&\checkmark&\textbf{63.14\%}&10.32cm& \textbf{82.94\%}\\
\hline
\end{tabular}}
\caption{Ablation studies on the different camera image sizes and camera network backbones. ResNet50 with 256x256 image size achieves the best performance in general.} %OKS@3D stands for OKS/ACC in 3D space. OKS@2D stands for OKS/ACC in 2D space. MPJPE is evaluated in 3D space.}
\label{tab:ablation_cam}
\end{table*}
\subsubsection{Ablation Study on Camera Image Size and Camera Network Backbone}
To study the effectiveness of modality fusion, experiments are conducted with different camera image sizes and camera network backbones with results in Table~\ref{tab:ablation_cam}. Here \emph{Inception 48x48} uses an Inception\cite{inception}-inspired convolutional network backbone with a 48x48 image size; \emph{Inception 64x64} is similar to \emph{Inception 48x48} but with a 64x64 image size; \emph{ResNet50 256x256} is the ResNet50 backbone used in the proposed method with a 256x256 image size. From the results in Table~\ref{tab:ablation_cam}, we observe that, even with smaller camera patch size and shallower backbone, the model still benefits from the additional camera modality. This observation is consistent with or without the auxiliary segmentation branch. With larger camera patch size and deeper backbone network, the overall performance is better.

We further studied the effect of different image sizes and network backbones on per-keypoint prediction errors in Table~\ref{tab:ablation_cam_per_keypoint}. These experiments are all with the proposed auxiliary segmentation branch. The results show that 1) Despite the choice of image size and backbone, addtional camera images generally bring considerable improvements on \emph{elbow}, \emph{wrist}, \emph{knee} and \emph{ankle}. This is because merely based on sparse and noisy LiDAR point clouds, accurately localizing these limb keypoints is difficult. Additional texture information from camera images makes the localization relatively easier. 2) Larger image size has better performance on most difficult keypoints like \emph{elbow} and \emph{wrist}. Surprisingly, it performs slightly worse than smaller patch sizes on other keypoints. 

Figure~\ref{fig:ablation} shows visualizations of 3D keypoint predictions on a pedestrian riding a scooter from the Waymo Open Dataset. It is a challenging case because of the objects (backpack, scooter) attached to the pedestrian and the irregular pose. The LiDAR-only model fails to predict accurate keypoints in Figure~\ref{fig:ablation_point_only}. By introducing modality fusion, improvements are observed on keypoints that are difficult to localize from sparse point clouds like those on the limbs (\emph{elbow}, \emph{wrist}, \emph{knee} and \emph{ankle}). The camera network used in the proposed method (ResNet50 on 256x256 images) predicts the most accurate keypoints (Figure~\ref{fig:ablation_r50}).
%The proposed camera network on 256x256 images in Figure~\ref{fig:ablation_r50} predicts the most accurate keypoints.

%-------------------------------------------------------------------------
\section{Conclusions}\label{sec:conclusion}
LiDAR based 3D HPE in AV differs from other applications for a variety of reasons including 3D resolution and range, absence of dense depth maps, and variation in test conditions. In this paper, we propose a multi-modal 3D HPE model with 2D weak supervision for autonomous driving. The model leverages both RGB camera images and LiDAR point clouds to tackle the challenges of 3D human pose estimation in unconstrained scenarios. Instead of using expensive 3D labels, the proposed model is trained on pure 2D labels. An auxiliary segmentation branch is added to introduce stronger supervision to the point network. Results on the Waymo Open Dataset (with evaluation labels to be released) and our internal dataset, and additional ablation studies showing the effectiveness of the proposed method.

{\small
\bibliographystyle{ieee_fullname}
\bibliography{macros,refs}
}

\end{document}

%%%%%%%%% TITLE
\title{Supplementary Material for Multi-Modal 3D Human Pose Estimation with 2D Weak Supervision in Autonomous Driving}

\author{Jingxiao Zheng\textsuperscript{1}
\and
Xinwei Shi\textsuperscript{1}

\and
Alexander Gorban\textsuperscript{1}

\and
Junhua Mao\textsuperscript{1}

\and
Yang Song\textsuperscript{1}

\and
Charles R. Qi\textsuperscript{1}

\and
Ting Liu\textsuperscript{2}

\and
Visesh Chari\textsuperscript{1}

\and
Andre Cornman\textsuperscript{1}

\and
Yin Zhou\textsuperscript{1}

\and
Congcong Li\textsuperscript{1} \quad Dragomir Anguelov\textsuperscript{1}
\\
\textsuperscript{1} Waymo LLC \quad\textsuperscript{2} Google Research\\
{\tt\small \{jingxiaozheng, xinweis, gorban, junhuamao, yangsong, rqi\}@waymo.com}, \\ {\tt\small liuti@google.com, \{visesh, cornman, yinzhou, congcongli, dragomir\}@waymo.com}
}

\maketitle

\section{Training Losses for Section 3.5}\label{sec:training_loss}
\textbf{Point Network:}
The training loss for the regression branch is a weighted Huber loss defined as
\begin{equation}
L_{\text{reg}}=\frac{1}{K}\sum_{k=1}^Kv_kr_kL_{\text{Huber}}\left(\frac{\hat{\mathbf{y}}_k^{(3)} - \tilde{\mathbf{y}}_k^{(3)}}{s_k}\right)
\end{equation}
% where $\hat{\mathbf{y}}_k^{(3)}$ is the 3D prediction from the model. $\tilde{\mathbf{y}}_k^{(3)}$ is the pseudo 3D label by label generation in Equation (1) in the original paper. $v_k$ is the visibility label of keypoint $k$ (0-1 valued). The loss is only applied on visible keypoints. $r_k$ is the label reliability. $s_k$ is the scaling factor of keypoint $k$. Keypoints that need to be more accurately localized will have smaller scales (therefore larger weights) during training.
where $\hat{\mathbf{y}}_k^{(3)}$ is the 3D prediction from the model, $\tilde{\mathbf{y}}_k^{(3)}$ is the pseudo 3D label by label generation in Equation (1) in the main paper, $v_k$ is the visibility label of keypoint $k$ (0-1 valued), $r_k$ is the label reliability (Section 3.4.1 in the main paper), and $s_k$ is the scaling factor of keypoint $k$. The loss is only applied on visible keypoints with $v_k$ being $1$. Keypoints that need to be more accurately localized have smaller scales $s_k$ (therefore larger weights) during training.

The loss for the segmentation branch is a weighted cross-entropy loss defined as
\begin{align}
L_{\text{seg}}=\frac{1}{K}\sum_{i=1}^N\sum_{k=1}^K&v_k\left\{w_{\text{pos}}l_{ik}\log p_{ik} + \right.\nonumber\\
&\left. w_{\text{neg}}(1-l_{ik})\log (1 - p_{ik})\right\}
\end{align}
%where $v_k$ is the visibility label of keypoint $k$. $w_{\text{pos}}$ and $w_{\text{neg}}$ are the weights that balance the positive and negative samples. $w_{\text{pos}}$ is usually much larger than $w_{\text{neg}}$ since for each keypoint there are much more points with negative labels than ones with positive labels.
where $v_k$ is the visibility label of keypoint $k$, and $w_{\text{pos}}$ and $w_{\text{neg}}$ are the weights balancing the positive and negative samples. $w_{\text{pos}}$ is usually much larger than $w_{\text{neg}}$ since for each keypoint there are much more points with negative labels than points with positive labels.

The overall loss for the point network is
\begin{equation}
L = L_{\text{reg}} + \lambda L_{\text{seg}}
\end{equation}
where $\lambda$ is used to weigh the auxiliary segmentation loss.

\noindent \textbf{Camera Network:} Similar to \cite{simple_baseline_2d}, the camera network is trained on a mean-squared-error loss with ground truth 2D heatmap as
\begin{equation}
L_{\text{cam}}=\frac{1}{H'W'K}\sum_{i,j=1}^{H',W'}\sum_{k=1}^Kv_k(h_{i,j,k}-g_{i,j,k})^2
\end{equation}
where $v_k$ is the visibility label of keypoint $k$, and $g_{i,j,k}$ is the ground truth heatmap generated by Gaussian functions centered at 2D ground truth keypoints. We train the camera network independently, then freeze it during point network training.

% \section{Labeling Details for Section 4.1}
% For 2D/3D keypoint labeling on the Waymo Open Dataset \cite{open_dataset} and our Internal Dataset, we adopt a definition of keypoints similar to the COCO Challenge \cite{coco}. Each keypoint is labeled by multiple labelers, whose results are aggregated to determine the final label. For 2D keypoint labeling, we only label 2D coordinates of keypoints that are visible in the camera image. For occluded keypoints, we label them as invisible. 3D labeling is similar, where we only label keypoints that are visible from the point clouds. Since the cameras have similar view angles to the main LiDAR, the occlusion status of keypoints is mostly consistent between 2D and 3D.

\begin{figure*}[t]
\centering
    \begin{subfigure}[b]{0.23\textwidth}
        \includegraphics[width=\textwidth]{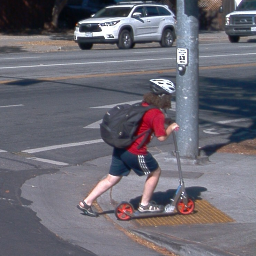}
        % \caption{Input camera image.} \label{fig:camera}
    \end{subfigure}
    ~
    \begin{subfigure}[b]{0.23\textwidth}
        \includegraphics[width=\textwidth]{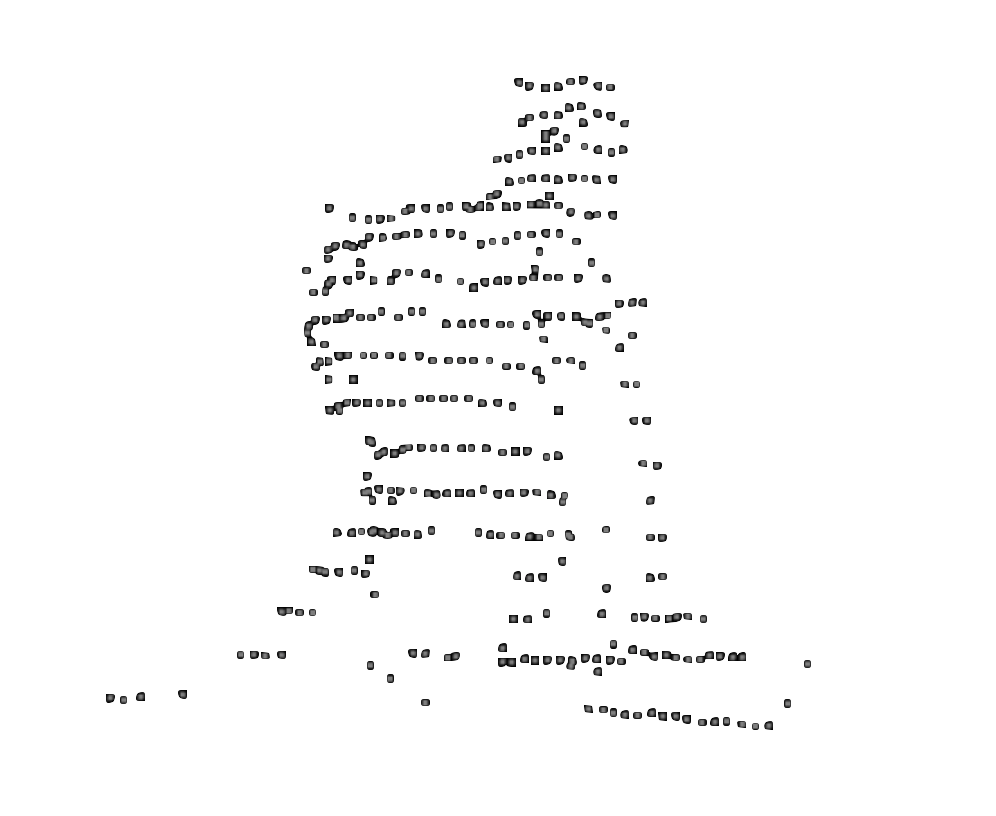}
        % \caption{Input LiDAR point cloud.} \label{fig:lidar}
    \end{subfigure}
    ~
    \begin{subfigure}[b]{0.23\textwidth}
        \includegraphics[width=\textwidth]{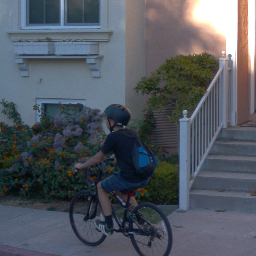}
        % \caption{Input LiDAR point cloud.} \label{fig:lidar}
    \end{subfigure}
    ~
    \begin{subfigure}[b]{0.23\textwidth}
        \includegraphics[width=\textwidth]{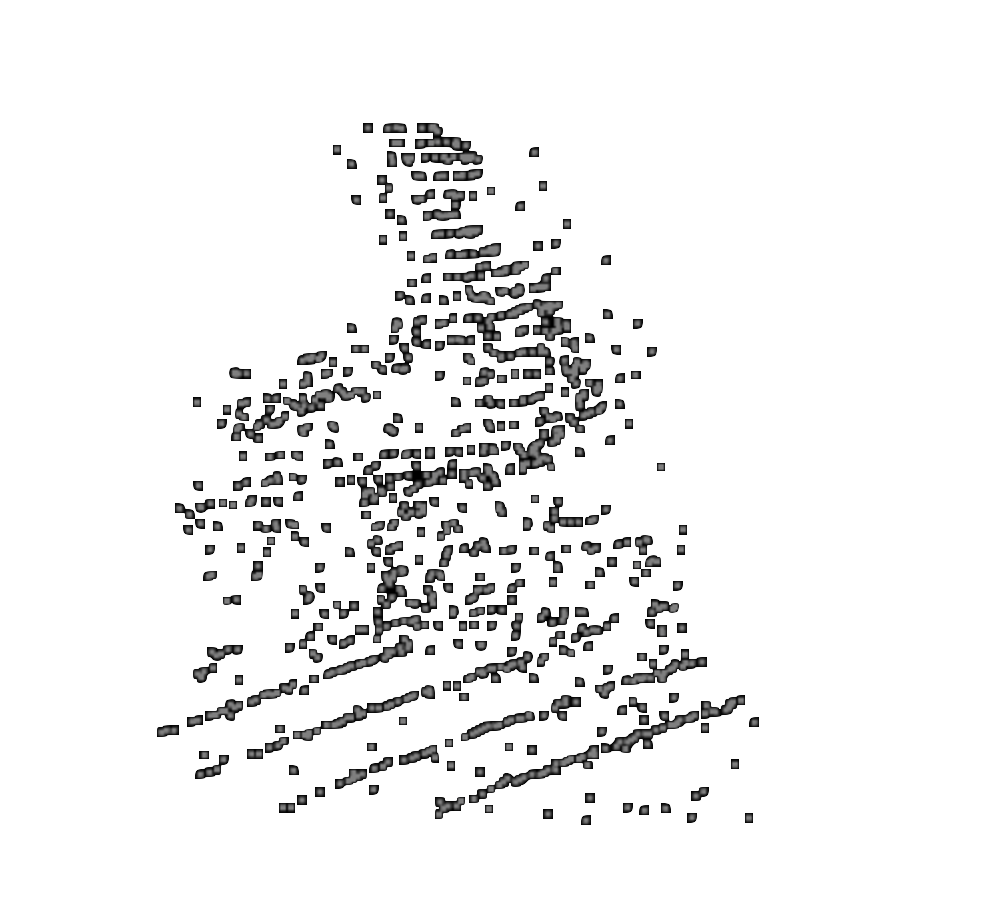}
        % \caption{Input LiDAR point cloud.} \label{fig:lidar}
    \end{subfigure}
    
    \begin{subfigure}[b]{0.23\textwidth}
        \includegraphics[width=\textwidth]{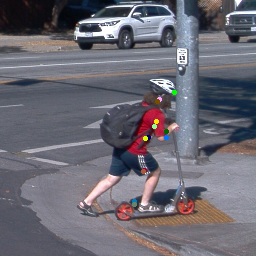}
        % \caption{LiDAR-only without segmentation branch.} \label{fig:point_baseline}
    \end{subfigure}
    ~
    \begin{subfigure}[b]{0.23\textwidth}
        \includegraphics[width=\textwidth]{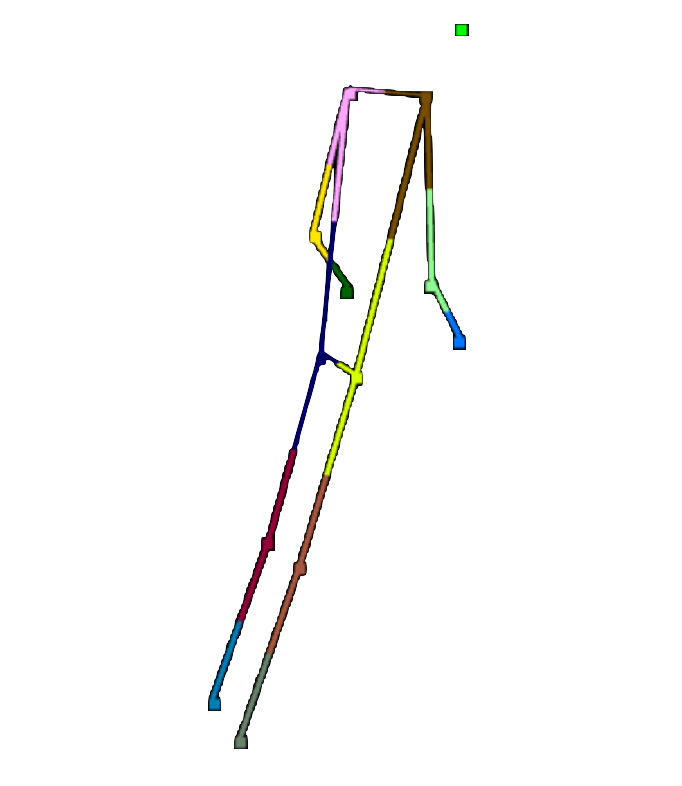}
        % \caption{LiDAR-only without segmentation branch.} \label{fig:point_baseline_3d}
    \end{subfigure}
    ~
    \begin{subfigure}[b]{0.23\textwidth}
        \includegraphics[width=\textwidth]{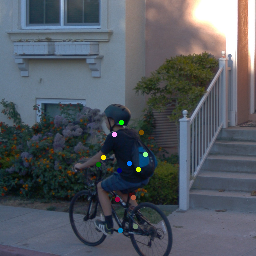}
        % \caption{Input LiDAR point cloud.} \label{fig:lidar}
    \end{subfigure}
    ~
    \begin{subfigure}[b]{0.23\textwidth}
        \includegraphics[width=\textwidth]{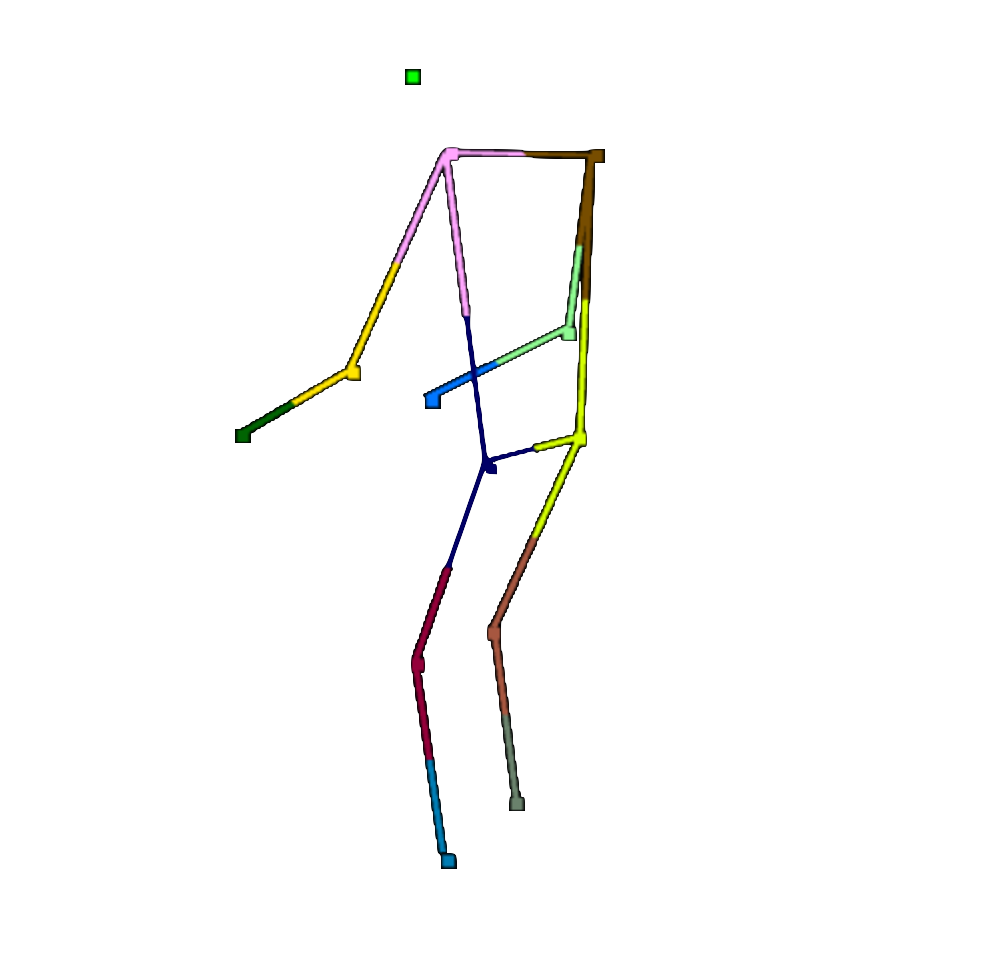}
        % \caption{Input LiDAR point cloud.} \label{fig:lidar}
    \end{subfigure}
    
    \begin{subfigure}[b]{0.23\textwidth}
        \includegraphics[width=\textwidth]{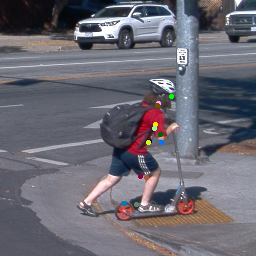}
        % \caption{LiDAR-only with segmentation branch.} \label{fig:point_segmentation}
    \end{subfigure}
    ~
    \begin{subfigure}[b]{0.23\textwidth}
        \includegraphics[width=\textwidth]{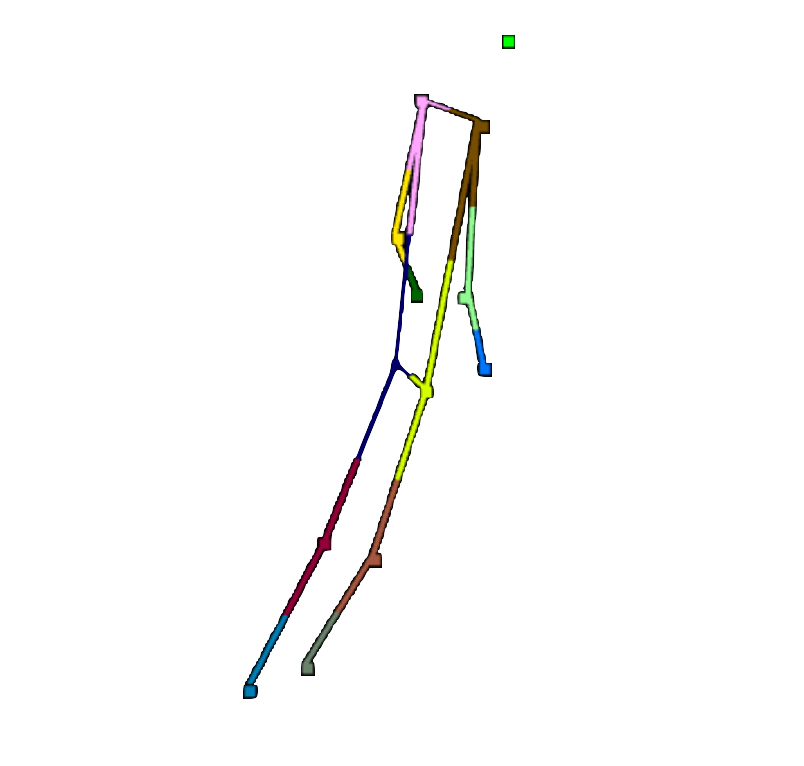}
        % \caption{LiDAR-only with segmentation branch.} \label{fig:point_segmentation_3d}
    \end{subfigure}
    ~
    \begin{subfigure}[b]{0.23\textwidth}
        \includegraphics[width=\textwidth]{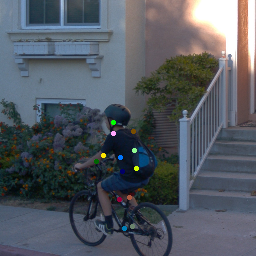}
        % \caption{Input LiDAR point cloud.} \label{fig:lidar}
    \end{subfigure}
    ~
    \begin{subfigure}[b]{0.23\textwidth}
        \includegraphics[width=\textwidth]{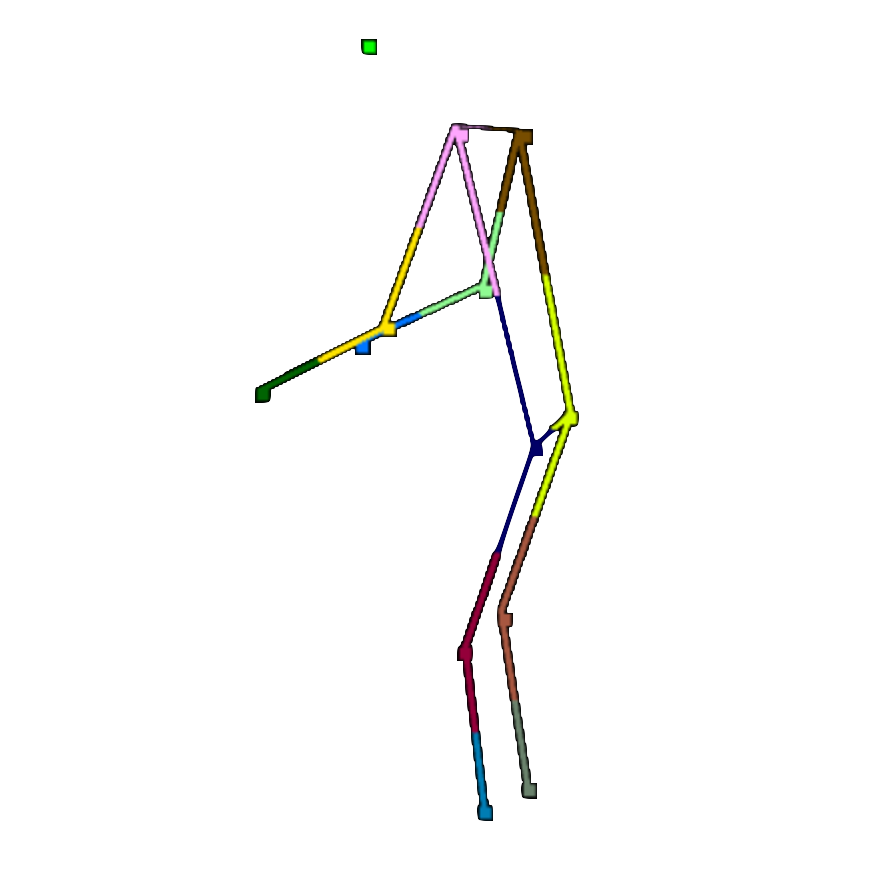}
        % \caption{Input LiDAR point cloud.} \label{fig:lidar}
    \end{subfigure}
    
    \begin{subfigure}[b]{0.23\textwidth}
        \includegraphics[width=\textwidth]{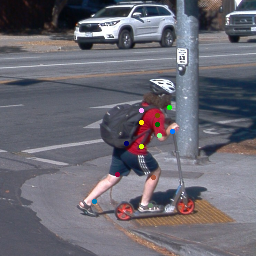}
        % \caption{Multi-modal without segmentation branch.} \label{fig:multimodal_noseg}
    \end{subfigure}
    ~
    \begin{subfigure}[b]{0.23\textwidth}
        \includegraphics[width=\textwidth]{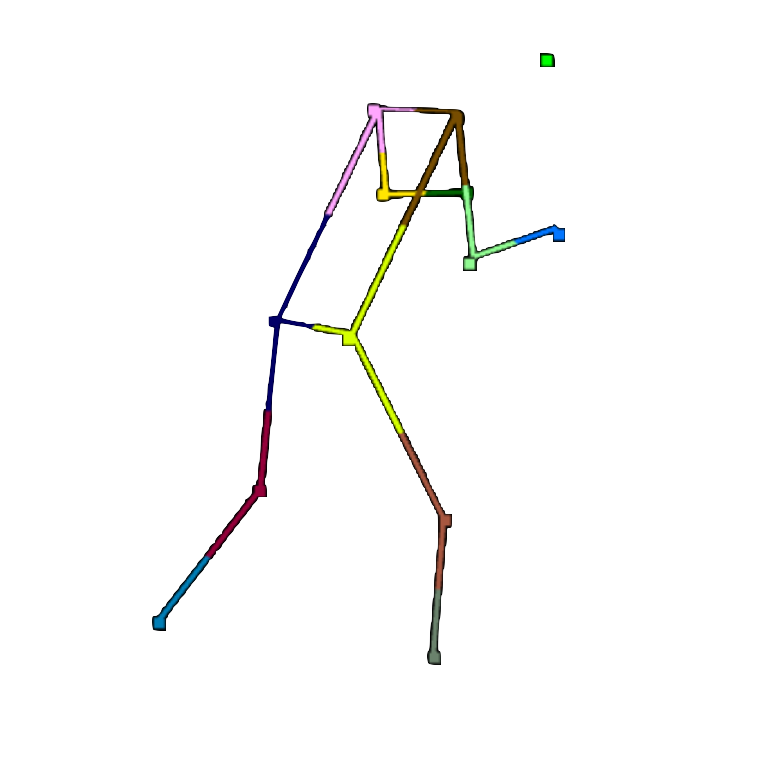}
        % \caption{Multi-modal without segmentation branch.} \label{fig:multimodal_noseg_3d}
    \end{subfigure}
    ~
    \begin{subfigure}[b]{0.23\textwidth}
        \includegraphics[width=\textwidth]{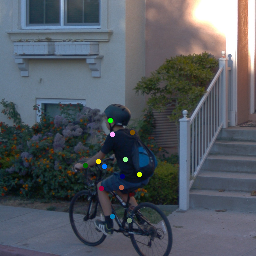}
        % \caption{Input LiDAR point cloud.} \label{fig:lidar}
    \end{subfigure}
    ~
    \begin{subfigure}[b]{0.23\textwidth}
        \includegraphics[width=\textwidth]{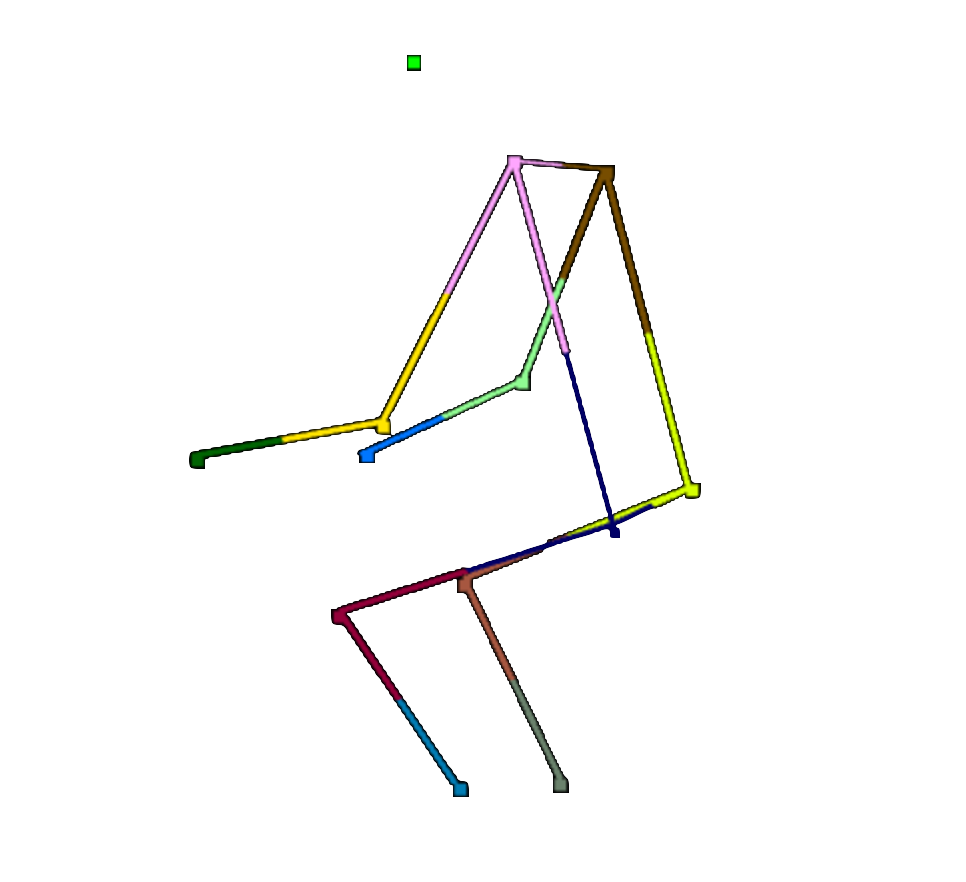}
        % \caption{Input LiDAR point cloud.} \label{fig:lidar}
    \end{subfigure}
    
    \begin{subfigure}[b]{0.23\textwidth}
        \includegraphics[width=\textwidth]{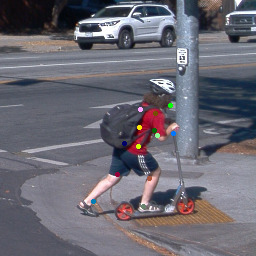}
        % \caption{Multi-modal with segmentation branch.} \label{fig:multimodal_segmentation}
    \end{subfigure}
    ~
    \begin{subfigure}[b]{0.23\textwidth}
        \includegraphics[width=\textwidth]{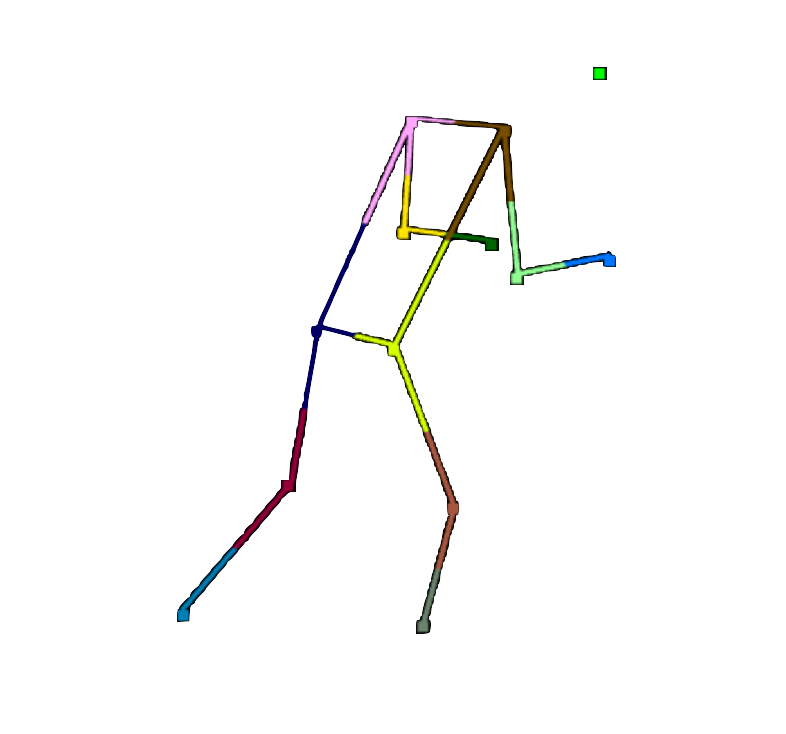}
        % \caption{Multi-modal with segmentation branch.} \label{fig:multimodal_segmentation_3d}
    \end{subfigure}
    ~
    \begin{subfigure}[b]{0.23\textwidth}
        \includegraphics[width=\textwidth]{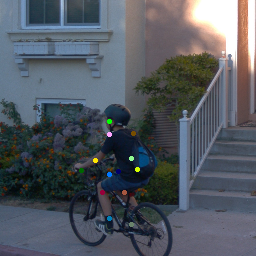}
        % \caption{Input LiDAR point cloud.} \label{fig:lidar}
    \end{subfigure}
    ~
    \begin{subfigure}[b]{0.23\textwidth}
        \includegraphics[width=\textwidth]{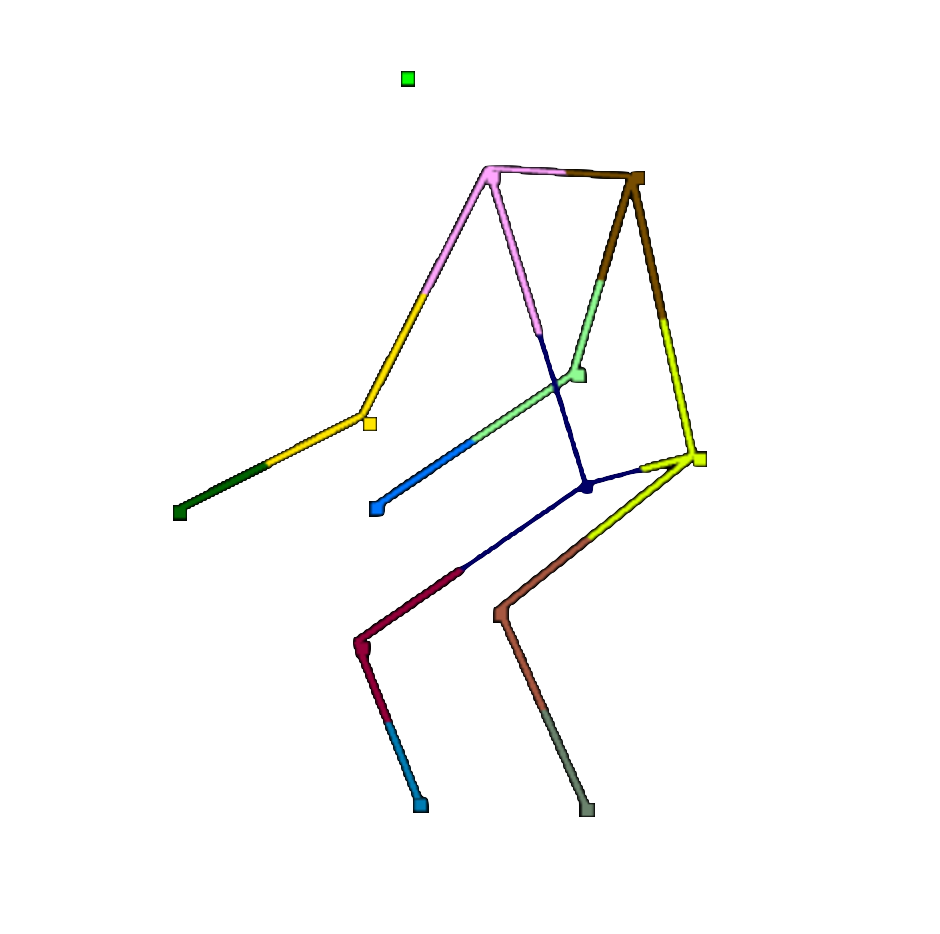}
        % \caption{Input LiDAR point cloud.} \label{fig:lidar}
    \end{subfigure}
\caption{Qualitative results from Waymo Open Dataset, comparing different model architectures similar to Table 3 in the main paper. Row 1 is the input camera image and LiDAR point cloud. Starting from Row 2, Columns 1 and 3 show the 2D projections of 3D predictions overlaid on camera images;  Columns 2 and 4 show 3D predictions. Row 2 to 5 correspond to LiDAR-only model without segmentation branch, LiDAR-only model with segmentation branch, multi-modal model without segmentation branch and multi-modal model with segmentation branch, respectively. Similar to the observations from Table 3 in the main paper, by adding key features to the model, the prediction accuracy improves consistently. Best viewed in color.}
\label{fig:example}
\end{figure*}
%Rows 2 to 5 correspond to rows in Table 3 in the main paper, which are LiDAR-only model without segmentation branch, LiDAR-only model with segmentation branch, multi-modal model without segmentation branch and multi-modal model with segmentation branch, respectively.

\begin{figure*}
\centering
\begin{tabular}{cccc}
 (a) LiDAR only & (b) LiDAR + Segmentation & (c) Multi-modal only & (d) Multi-modal + Segmentation \\
\begin{subfigure}[b]{0.22\textwidth} \includegraphics[width=\textwidth]{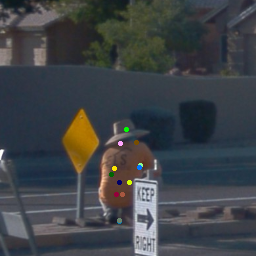}
\end{subfigure} &
\begin{subfigure}[b]{0.22\textwidth} \includegraphics[width=\textwidth]{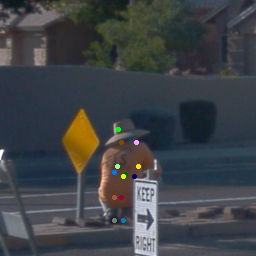}
\end{subfigure} &
\begin{subfigure}[b]{0.22\textwidth} \includegraphics[width=\textwidth]{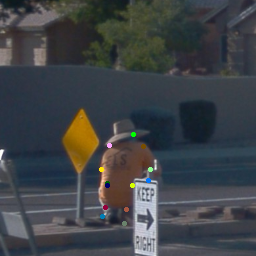}
\end{subfigure} &
\begin{subfigure}[b]{0.22\textwidth} \includegraphics[width=\textwidth]{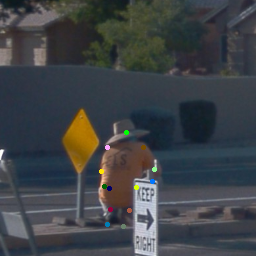}
\end{subfigure} \\
\begin{subfigure}[b]{0.22\textwidth} \includegraphics[width=\textwidth]{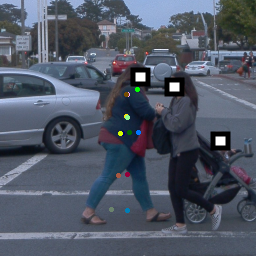}
\end{subfigure} &
\begin{subfigure}[b]{0.22\textwidth} \includegraphics[width=\textwidth]{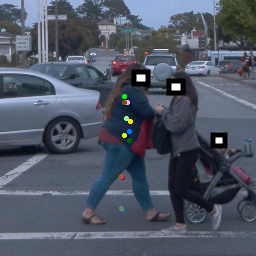}
\end{subfigure} &
\begin{subfigure}[b]{0.22\textwidth} \includegraphics[width=\textwidth]{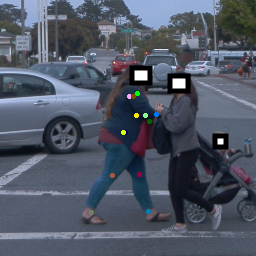}
\end{subfigure} &
\begin{subfigure}[b]{0.22\textwidth} \includegraphics[width=\textwidth]{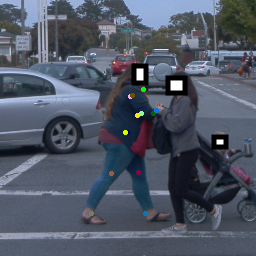}
\end{subfigure}\\
\begin{subfigure}[b]{0.22\textwidth} \includegraphics[width=\textwidth]{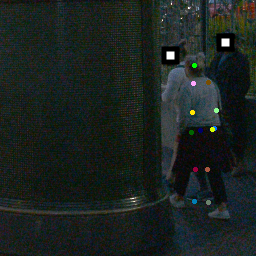}
\end{subfigure} &
\begin{subfigure}[b]{0.22\textwidth} \includegraphics[width=\textwidth]{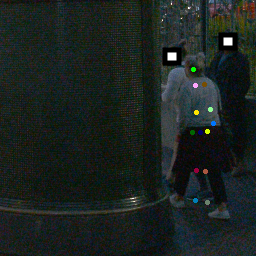}
\end{subfigure} &
\begin{subfigure}[b]{0.22\textwidth} \includegraphics[width=\textwidth]{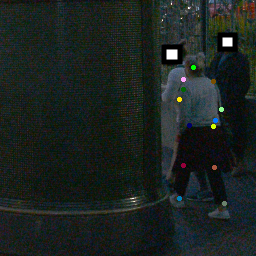}
\end{subfigure} &
\begin{subfigure}[b]{0.22\textwidth} \includegraphics[width=\textwidth]{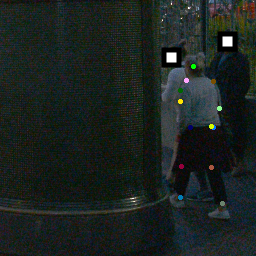}
\end{subfigure} \\
\begin{subfigure}[b]{0.22\textwidth} \includegraphics[width=\textwidth]{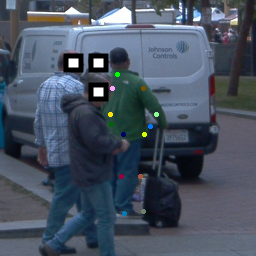}
\end{subfigure} &
\begin{subfigure}[b]{0.22\textwidth} \includegraphics[width=\textwidth]{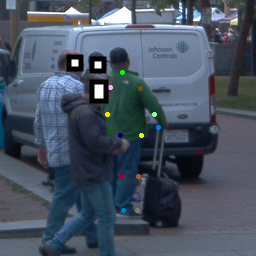}
\end{subfigure} &
\begin{subfigure}[b]{0.22\textwidth} \includegraphics[width=\textwidth]{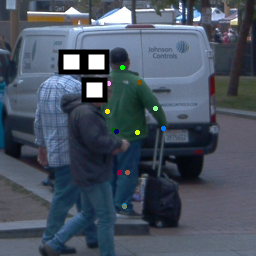}
\end{subfigure} &
\begin{subfigure}[b]{0.22\textwidth} \includegraphics[width=\textwidth]{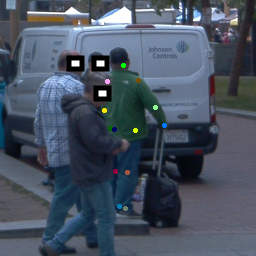}
\end{subfigure}\\
\begin{subfigure}[b]{0.22\textwidth} \includegraphics[width=\textwidth]{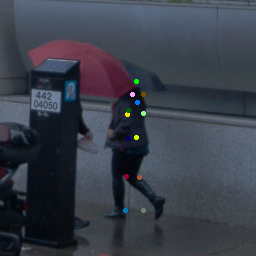}
\end{subfigure} &
\begin{subfigure}[b]{0.22\textwidth} \includegraphics[width=\textwidth]{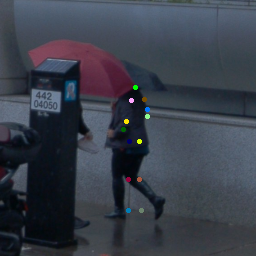}
\end{subfigure} &
\begin{subfigure}[b]{0.22\textwidth} \includegraphics[width=\textwidth]{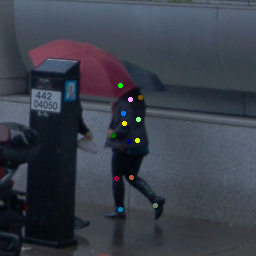}
\end{subfigure} &
\begin{subfigure}[b]{0.22\textwidth} \includegraphics[width=\textwidth]{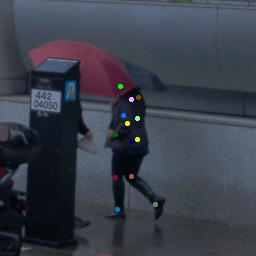}
\end{subfigure}
\end{tabular}
\caption{Additional qualitative results on the Waymo Open Dataset, showing the improvement that our approach brings over LiDAR-only model. The columns in each row show: a) LiDAR-only model without segmentation branch; b) LiDAR-only model with segmentation branch; c) multi-modal model without segmentation branch; and d) multi-modal model with segmentation branch. Note that in each case there is either self- or other forms of occlusion that deteriorates LiDAR only results. While segmentation and camera each can provide some additional clue, combining everything produces the best result. Best viewed in color.}
\label{fig:example2}
\end{figure*}

\begin{figure*}
\begin{tabular}{cccc}
 (a) LiDAR only & (b) LiDAR + Segmentation & (c) Multi-modal only & (d) Multi-modal + Segmentation \\
\begin{subfigure}[b]{0.22\textwidth} \includegraphics[width=\textwidth]{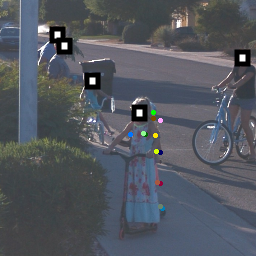}
\end{subfigure} &
\begin{subfigure}[b]{0.22\textwidth} \includegraphics[width=\textwidth]{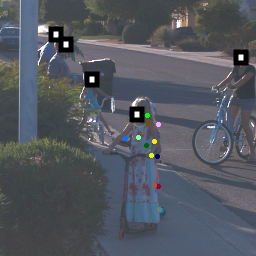}
\end{subfigure} &
\begin{subfigure}[b]{0.22\textwidth} \includegraphics[width=\textwidth]{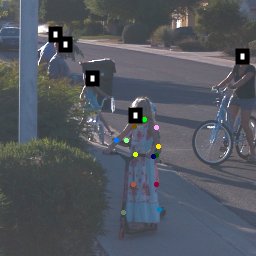}
\end{subfigure} &
\begin{subfigure}[b]{0.22\textwidth} \includegraphics[width=\textwidth]{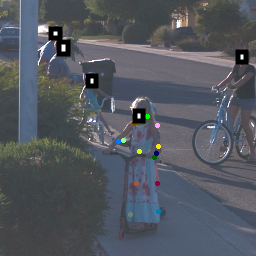}
\end{subfigure} \\
\begin{subfigure}[b]{0.22\textwidth} \includegraphics[width=\textwidth]{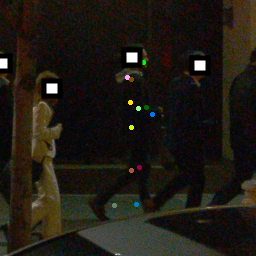}
\end{subfigure} &
\begin{subfigure}[b]{0.22\textwidth} \includegraphics[width=\textwidth]{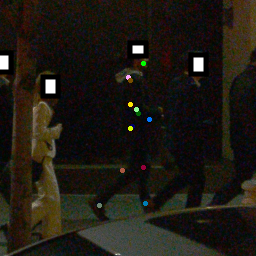}
\end{subfigure} &
\begin{subfigure}[b]{0.22\textwidth} \includegraphics[width=\textwidth]{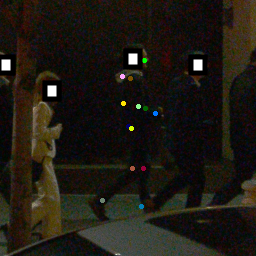}
\end{subfigure} &
\begin{subfigure}[b]{0.22\textwidth} \includegraphics[width=\textwidth]{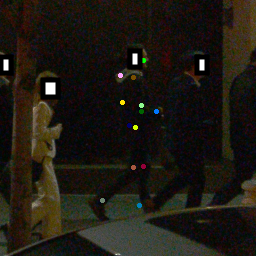}
\end{subfigure} \\
\begin{subfigure}[b]{0.22\textwidth} \includegraphics[width=\textwidth]{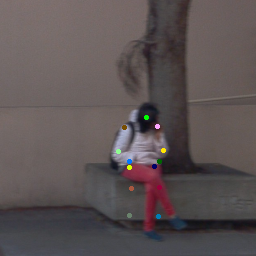}
\end{subfigure} &
\begin{subfigure}[b]{0.22\textwidth} \includegraphics[width=\textwidth]{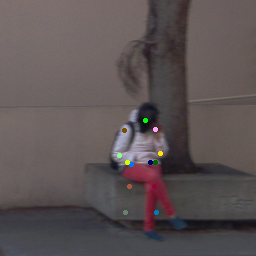}
\end{subfigure} &
\begin{subfigure}[b]{0.22\textwidth} \includegraphics[width=\textwidth]{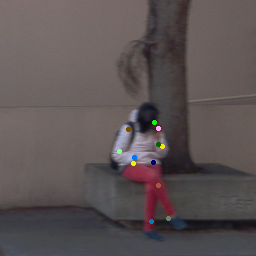}
\end{subfigure} &
\begin{subfigure}[b]{0.22\textwidth} \includegraphics[width=\textwidth]{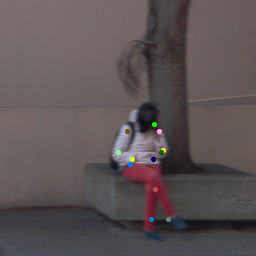}
\end{subfigure} \\
\begin{subfigure}[b]{0.22\textwidth} \includegraphics[width=\textwidth]{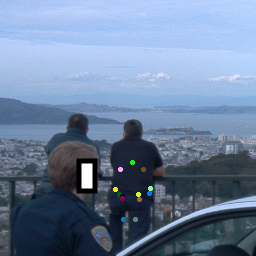}
\end{subfigure} &
\begin{subfigure}[b]{0.22\textwidth} \includegraphics[width=\textwidth]{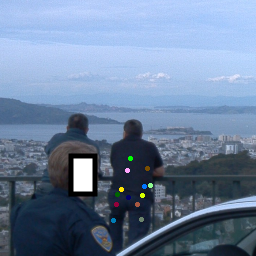}
\end{subfigure} &
\begin{subfigure}[b]{0.22\textwidth} \includegraphics[width=\textwidth]{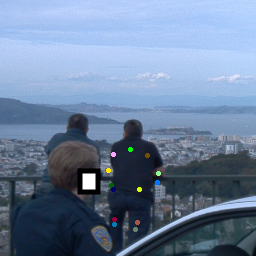}
\end{subfigure} &
\begin{subfigure}[b]{0.22\textwidth} \includegraphics[width=\textwidth]{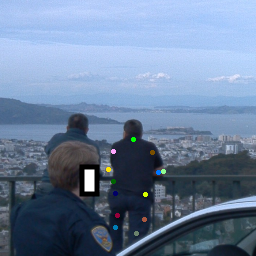}
\end{subfigure} \\
\begin{subfigure}[b]{0.22\textwidth} \includegraphics[width=\textwidth]{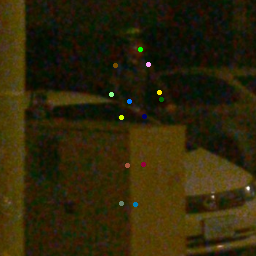}
\end{subfigure} &
\begin{subfigure}[b]{0.22\textwidth} \includegraphics[width=\textwidth]{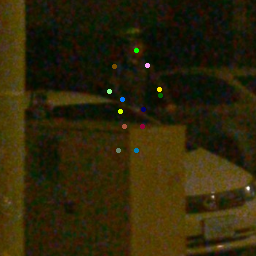}
\end{subfigure} &
\begin{subfigure}[b]{0.22\textwidth} \includegraphics[width=\textwidth]{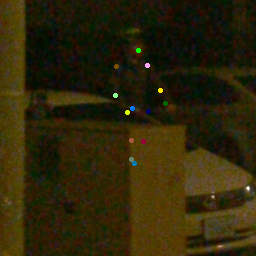}
\end{subfigure} &
\begin{subfigure}[b]{0.22\textwidth} \includegraphics[width=\textwidth]{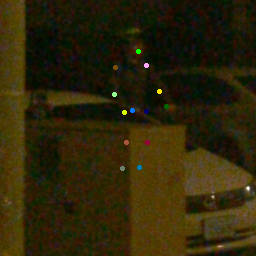}
\end{subfigure}
\end{tabular}
%\caption{In qualitative results, we notice that movement along with self occlusion plays a crucial role in highlighting differences between the various approaches. Note above that either the persons are self occluded, or the vehicle is in motion during the event, or both. Even when the human is extremely occluded (last row), our approach helps us learn reasonable priors for occluded keypoints (in this case, riding a bike)}
\caption{More qualitative results from the Waymo Open Dataset, highlighting differences between various architectures. Note that even when the human is heavily occluded (last row), our approach can get reasonable results from the learned priors (in this case, riding a bike). Best viewed in color.}
\label{fig:example3}
\end{figure*}

\section{Metrics for Section 4.1}
\subsection{OKS/ACC Metric}
%In this paper we focus on pose estimation instead of keypoint detection by assuming that we have already successfully detected the person and have exactly one estimated pose for each ground-truth pose. Therefore instead of using the OKS/AP metric defined in COCO keypoint challenge, we introduce a similar OKS/ACC metric for evaluation:
This paper focuses on pose estimation instead of keypoint detection by assuming that the person has been successfully detected and there is exactly one estimated pose for each ground-truth pose. Therefore, instead of using the OKS/AP metric defined in COCO keypoint challenge \cite{coco_keypoint_evaluation}, we introduce a modified OKS/ACC metric for evaluation:
\begin{equation}
ACC^{OKS=t} = \frac{1}{N}\sum_{n=1}^N\mathbbm{1}\left\{\mathrm{OKS}_n \geq t\right\}
\end{equation}
where $t$ is the threshold on OKS, $N$ is the total number of samples in the test set, and $\mathrm{OKS}_n$ is the OKS of prediction on sample $n$. In our experiments we averaged OKS/ACC over $t$ from $0.5$ to $0.95$ with a step-size of $0.05$.
%$t = 0.5:0.05:0.95$.
\subsection{Per-keypoint OKS}
%Following the OKS metric defined in COCO keypoint challenge, we compute per-keypoint OKS as
Per-keypoint OKS is defined as OKS \cite{coco_keypoint_evaluation} for one keypoint type. For keypoint type $i$, 
\begin{equation}
    OKS = \frac{\exp(-d_i^2/2s^2k_i^2)\delta(v_i>0)}{\delta(v_i>0) + \epsilon}
\end{equation}
where $d_i$ is the distance between ground truth and prediction, $v_i$ is visibility of the ground truth, $s$ is the object scale, $k_i$ is a per-keypoint constant, and $\epsilon$ is a small number here to prevent zero denominator.

\section{Implementation Details for Section 4.1}
%The proposed model is trained on the Internal Dataset. Evaluation is performed on the Waymo Open Dataset and the test partition of the Internal Dataset. The training and testing configurations for both datasets are the same. 
The following are some details on training.
For the camera network, we resize all input images to $256\times256$. The output heatmap size is $64\times64\times13$ (13 keypoints are predicted). 
%We train it on the camera images of pedestrians after the internal training set with an Adam optimizer and batch size $32\times32$ for 40000 iterations. 
The camera network is trained with an Adam optimizer and batch size $32\times32$ for 40000 iterations. The initial learning rate is $1\times 10^{-4}$ and is decayed by $0.1$ at 20000 and 30000 iterations. Random augmentation is applied during training, so each input image is randomly rotated, scaled or flipped. The heatmap is furthered smoothed by a $7\times7$ Gaussian kernel with $\sigma=3$.

For the point network, we sub-sample the input point cloud to a fixed size of 256 points. We only use the 3D coordinates of points as point feature, which is concatenated with the 13-dimensional camera feature from the camera network to perform modality fusion. We set $\lambda=0.1$ for the segmentation task (Equation (3) in the main paper). The network is trained
%on the point clouds from the internal training set together with the camera features 
for 100000 iterations, with an SGD optimizer and batch size $128$. The initial learning rate is $1\times10^{-3}$ and is decayed by cosine decay. During training, input point clouds are rotated in the X-Y plane by a random angle in $[0, 2\pi)$ as data augmentation.

\section{Qualitative Results}
Figures~\ref{fig:example}, ~\ref{fig:example2} , and ~\ref{fig:example3} show qualitative results from the Waymo Open Dataset.
%Figure~\ref{fig:example} shows qualitative results from the Waymo Open Dataset, 
Figure~\ref{fig:example} compares different model architectures corresponding to Table 3 in the main paper. Row 1 shows the input camera image and LiDAR point cloud. Starting from Row 2, Columns 1 and 3 show the 2D projections of 3D predictions overlaid on camera images, and Columns 2 and 4 show 3D predictions.
Rows 2 to 5 correspond to rows in Table 3 in the main paper, which are LiDAR-only model without segmentation branch, LiDAR-only model with segmentation branch, multi-modal model without segmentation branch and multi-modal model with segmentation branch, respectively.

Due to the objects (e.g., backpack, scooter, bike) attached to the pedestrian and the pose of the legs, the input LiDAR point cloud looks different from a regular pedestrian, which poses challenges for LiDAR-only 3D HPE. From the results, we can see that it is difficult to predict accurate keypoints (especially lower body keypoints) from LiDAR point cloud only (Rows 2 and 3). By utilizing texture information from the camera image, multi-modal architectures show much better performance (Rows 4 and 5) on all keypoints. On the other hand, comparing Rows 2, 4 with Rows 3, 5 respectively, we see that adding segmentation branch refines the predictions for both LiDAR-only and multi-modal architectures. Similar to the observations from Table 3 in the main paper, by adding key features to the model, the prediction accuracy improves consistently.

%From these qualitative results, we observe the similar trend to Table 3: by adding key features to the model, the prediction accuracy improves consistently.

%\caption{Qualitative results from Waymo Open Dataset, comparing different model architectures similar to Table 3 in the main paper. Row 1 is the input camera image and LiDAR point cloud. Starting from Row 2, Columns 1 and 3 show the 2D projections of 3D predictions overlaid on camera images;  Columns 2 and 4 show 3D predictions. Row 2 to 5 correspond to LiDAR-only model without segmentation branch, LiDAR-only model with segmentation branch, multi-modal model without segmentation branch and multi-modal model with segmentation branch, respectively. Similar to the observations from Table 3 in the main paper, by adding key features to the model, the prediction accuracy improves consistently. Best viewed in color.}

Figure~\ref{fig:example2} gives additional qualitative results in cases of occlusion. In each of these cases, we can see how adding camera information to LiDAR provides a big boost, especially for identifying the individual limbs of the subject in question. This is understandable, since with occlusion, it is often difficult to isolate LiDAR point clouds of a person from their immediate surrounding. Figure~\ref{fig:example3} shows more qualitative results, highlighting differences between various architectures. Note that even when the human is heavily occluded (last row), our approach can get reasonable results from the learned priors.

\label{fig:example2}

%\caption{More qualitative results from the Waymo Open Dataset, highlighting differences between various architectures. Note that even when the human is heavily occluded (last row), our approach can get reasonable results from the learned priors (in this case, riding a bike).}

{\small
\bibliographystyle{ieee_fullname}
\bibliography{refs}
}